\documentclass{article}
\usepackage[preprint]{neurips_2026}
\usepackage[utf8]{inputenc}
\usepackage[T1]{fontenc}
\usepackage{hyperref}
\usepackage{url}
\usepackage{booktabs}
\usepackage{amsfonts}
\usepackage{nicefrac}
\usepackage{microtype}
\usepackage{enumitem}
\usepackage{amsmath}
\usepackage{amsthm}
\usepackage{amssymb}
\usepackage{multirow}
\usepackage{makecell}
\usepackage{graphicx}
\usepackage{subcaption}
\usepackage{wrapfig}
\usepackage{caption}
\usepackage{colortbl}
\usepackage{algorithm}
\usepackage{algorithmic}
\usepackage{tikz}
\usetikzlibrary{arrows.meta, positioning, shapes.geometric, fit, backgrounds}

\usepackage[most]{tcolorbox}
\usepackage{xcolor}
\usepackage{varwidth}
\usepackage{listings}
\definecolor{FMBBlue}{HTML}{2F5F8F}
\definecolor{FMBLightBlue}{HTML}{F3F8FC}
\definecolor{FMBBorder}{HTML}{A9C7E2}
\definecolor{FMBGray}{HTML}{4A5568}

\definecolor{HMBlue}{HTML}{0B55A0}
\definecolor{HMSoftBlue}{HTML}{F5F9FF}
\definecolor{HMBorderBlue}{HTML}{8CB7E8}
\definecolor{HMRed}{HTML}{B91C1C}
\definecolor{HMSoftRed}{HTML}{FFF5F5}
\definecolor{HMGray}{HTML}{4A5568}
\definecolor{HMCodeBg}{HTML}{F7F8FA}
\definecolor{HMCodeBorder}{HTML}{D6D9DE}

\newtcolorbox{formalbenchproblem}[2]{
  enhanced,
  breakable,
  colback=FMBLightBlue,
  colframe=FMBBorder,
  boxrule=0.6pt,
  arc=2mm,
  left=9pt,
  right=9pt,
  top=8pt,
  bottom=8pt,
  before skip=0.9em,
  after skip=1.1em,
  title={
    \sffamily\bfseries\color{FMBBlue} #1
    \hfill
    {\normalfont\ttfamily\small\color{FMBGray} #2}
  },
  colbacktitle=white,
  coltitle=FMBBlue,
  boxed title style={
    colback=white,
    colframe=white,
    boxrule=0pt
  },
  attach boxed title to top left={
    xshift=7pt,
    yshift=-2pt
  },
  varwidth boxed title,
  fontupper=\small,
  parbox=false
}

\newcommand{\problemstatement}{\par\smallskip\noindent\textbf{Statement.}\quad}
\newcommand{\problemtarget}{\par\smallskip\noindent\textbf{Target.}\quad}

\lstdefinestyle{hardmathlean}{
  basicstyle=\ttfamily\footnotesize,
  columns=fullflexible,
  keepspaces=true,
  breaklines=true,
  breakatwhitespace=false,
  showstringspaces=false,
  frame=none,
  aboveskip=0pt,
  belowskip=0pt,
  literate={ℝ}{{$\mathbb{R}$}}1
           {→}{{$\to$}}1
           {λ}{{$\lambda$}}1
           {∀}{{$\forall$}}1
           {∃}{{$\exists$}}1
           {ε}{{$\varepsilon$}}1
           {∫}{{$\int$}}1
}

\newtcolorbox{hardmathbluebox}[1]{
  enhanced,
  breakable,
  colback=HMSoftBlue,
  colframe=HMBorderBlue,
  colbacktitle=HMBlue,
  coltitle=white,
  fonttitle=\sffamily\bfseries\small,
  title={#1},
  boxrule=0.6pt,
  arc=2mm,
  left=10pt,
  right=10pt,
  top=8pt,
  bottom=8pt,
  before skip=0.9em,
  after skip=1.1em,
  fontupper=\small
}

\newtcolorbox{hardmathredbox}[1]{
  enhanced,
  breakable,
  colback=HMSoftRed,
  colframe=HMRed,
  colbacktitle=HMRed,
  coltitle=white,
  fonttitle=\sffamily\bfseries\small,
  title={#1},
  boxrule=0.6pt,
  arc=2mm,
  left=10pt,
  right=10pt,
  top=8pt,
  bottom=8pt,
  before skip=0.9em,
  after skip=1.1em,
  fontupper=\small
}

\newtcolorbox{hardmathcodebox}[1]{
  enhanced,
  breakable,
  colback=HMCodeBg,
  colframe=HMCodeBorder,
  colbacktitle=HMCodeBorder!45,
  coltitle=black,
  fonttitle=\sffamily\bfseries\small,
  title={#1},
  boxrule=0.5pt,
  arc=1.5mm,
  left=8pt,
  right=8pt,
  top=7pt,
  bottom=7pt,
  before skip=0.8em,
  after skip=1.0em
}

\newcommand{\hardmathmodelheading}[1]{%
  \par\medskip
  {\large\sffamily\bfseries\color{HMBlue}#1}\par
  \vspace{0.25em}{\color{HMBorderBlue}\hrule height 0.6pt}\vspace{0.75em}
}

\newcommand{\hardmathsmalllabel}[1]{\par\smallskip\noindent\textbf{#1}\quad}

\theoremstyle{remark}

\title{MathCoPilot: An Interactive System for Human--AI Symbiotic Paradigm of Mathematical Research}

\author{
\parbox{\linewidth}{\centering \bfseries
  Junjie Zhang\textsuperscript{1*}
  \  Jiayu Liu\textsuperscript{1*}
  \  Wenbin Liu\textsuperscript{1*}
  \  Zhenya Huang\textsuperscript{1$\dag$}
  \  Doudou Wang\textsuperscript{1,2}
  \  Yan Jiang\textsuperscript{1,2}
  \  Leiye Xu\textsuperscript{1,2}
 \  Tao Xiong\textsuperscript{1,2}
 \  Wen Huang\textsuperscript{1,2}
  \  Qi Liu\textsuperscript{1}
  \  Guoping Hu\textsuperscript{1}
  \  Enhong Chen\textsuperscript{1}
  \  Mengping Zhang\textsuperscript{1,2$\dag$}
  \  Xiangdong Ye\textsuperscript{1,2}
  }
  \\ 
  \vspace{-0.1cm} 
  \\
 \textsuperscript{1}State Key Laboratory of Cognitive Intelligence, University of Science and Technology of China \\
 \textsuperscript{2}School of Mathematical Sciences, University of Science and Technology of China \\
  {\normalfont\small \textsuperscript{*}Equal contribution \quad \textsuperscript{\dag}Corresponding authors} \\
  \vspace{0.15cm}
  {\normalfont\textbf{Webpage:} \sysurl{}}
}

\newcommand{\sysname}{\textsc{MathCoPilot}}
\newcommand{\sysurl}{\href{https://mathcopilot.cn/}{\color{HMBlue}\texttt{https://mathcopilot.cn}}}
\newcommand{\heatZero}[1]{\cellcolor{red!6}#1}
\newcommand{\heatLow}[1]{\cellcolor{yellow!18}#1}
\newcommand{\heatMid}[1]{\cellcolor{green!18}#1}
\newcommand{\heatHigh}[1]{\cellcolor{green!35}#1}
\newcommand{\heatFull}[1]{\cellcolor{green!55}#1}
\newcommand{\passmark}{\textcolor{green!60!black}{\textbf{Pass}}}
\newcommand{\failmark}{\textcolor{red}{\textbf{Fail}}}

\begin{document}

\maketitle

\begin{abstract}
Existing LLM-based theorem provers have achieved impressive results on formal mathematics benchmarks, yet they remain confined to acting as autonomous agents that prove a stated proposition. In this paper, we propose \sysname{}, a human-in-the-loop system that embodies a new human--AI symbiotic paradigm for mathematical research, in which the mathematician steers the high-level mathematical direction while AI agents carry out the detailed formalization and proof work under continuous human guidance. \sysname{} unifies three core capabilities: \emph{(1)} an interactive workbench where the mathematician and AI agents collaborate through a living proof blueprint that decomposes a proof into navigable steps the human can directly inspect, direct, and refine; \emph{(2)} automated proving skill orchestration with adaptive knowledge base search and Lean-integrated iterative verification; and \emph{(3)} topic-driven paper retrieval and automated formalization into a verified Lean knowledge base. Using \sysname{}, we systematically compare four state-of-the-art LLMs, including Gemini~3.1~Pro, GPT-5.4, and Claude~Opus~4.7, on a FormalMATH subset and on two real PDE theorems requiring deep domain expertise, evaluating their ability to produce verified Lean~4 proofs and to identify errors in deliberately incorrect proofs. Our results show that while current models can handle undergraduate-level problems with high success rates under favorable autoformalization conditions, substantial challenges remain for domain-specific theorems requiring genuine mathematical understanding. 
\end{abstract}

\section{Introduction}\label{sec:intro}

Mathematical research is undergoing a paradigm shift~\citep{davies2021advancing,tao2025machine,sothanaphan2026resolution}. For most of its history, the machine has been at most a tool, a calculator, a search engine, or a proof checker, while the mathematician alone poses the questions, devises the arguments, and judges their correctness. The rise of capable AI agents is now challenging this division of labor and pointing toward a new paradigm in which AI and mathematician work as collaborators, the human steering the high-level direction and creation while the machine carries out ever larger parts of the reasoning. The dream of such assistance dates back to the origins of automated theorem proving \citep{irving2016deepmath}, and the emergence of large language models (LLMs) has accelerated it, with systems such as AlphaProof \citep{hubert2025olympiad}, DeepSeek-Prover \citep{xin2024deepseek}, and various Lean-based agents \citep{jiangdraft, poluformal, liu2026numina,ju2026automated} achieving impressive results. However, \emph{these systems are inherently task-centered, concentrating on how an agent (or multi agents) can best carry out one concrete step such as proving a given theorem, while overlooking the mathematician's role in research and failing to support their broader workflow.}

The workflow of a working mathematician is far more complex than ``prove a given theorem.'' It involves several interconnected activities:
\textbf{Reading and digesting existing work.} Mathematicians spend a substantial fraction of their time understanding papers, proofs, and constructions produced by others.
\textbf{Long-term knowledge accumulation.} Each researcher maintains a personal repository of lemmas, notation conventions, and proof techniques built over years of study.
\textbf{Interactive human-machine collaboration.} Researchers need to ask questions, discuss approaches, request explanations, and have their informal sketches formalized by an assistant that adapts to their style.
\textbf{Diverse proof strategies.} Mathematicians think in heterogeneous ways: some begin with natural language sketches, while others prefer to formalize first and then reason, so a useful assistant must support multiple strategies rather than committing to a single approach.

Current LLM-based formal mathematics systems primarily operate under a \emph{question-and-answer (Q\&A) paradigm}~\citep{liu2026numina,baba2025prover,lin2025goedel,kim2025process}. While they are capable of generating complete proof processes, this black-box approach falls short of genuinely assisting mathematicians in active research. First, because these systems directly output complete, end-to-end proofs in a single pass, the generated content is often overwhelming. Mathematicians are forced to parse a massive volume of unsegmented steps, making it difficult to read through the proof, comprehend the underlying logic, or verify its correctness. Second, these systems lack interactivity across levels of granularity. Whether a mathematician wants to propose a high-level proof sketch or intervene at a specific logical step, the Q\&A format prevents them from injecting insights or steering the model's direction mid-generation. Due to these rigidities, practical research tasks remain fundamentally unsupported. A mathematician who wants to formalize a proof from a newly published paper, explore whether a lemma from an earlier result applies, or discuss a proof strategy with an AI collaborator currently has no unified tool to turn to. Moreover, the rapid proliferation of LLM-based provers has made it difficult to understand the relative strengths of different models and strategies. Existing evaluations typically report a single pass-rate number per system, obscuring important questions: how does performance vary across proving strategies? How sensitive are results to the choice of autoformalization pipeline? Can current models detect subtle logical errors in proofs?

To address these gaps, we present \sysname{}, a human-in-the-loop system that embodies a new \emph{human--AI symbiotic paradigm} for mathematical research. Rather than treating proving as an autonomous task, \sysname{} casts it as a collaboration in which the mathematician steers the high-level mathematical direction while AI agents carry out the detailed implementation. Given a research topic, the system supports paper retrieval and automatically transforms the retrieved content into verified Lean code stored in a local knowledge base. When proving theorems, it orchestrates multiple proving skills, dynamically switching between them based on intermediate progress, and automatically verifies each candidate proof in Lean. Crucially, instead of having AI agents converse among themselves, \sysname{} keeps the mathematician in control through a living proof blueprint, a flowchart of proof steps in which the human provides direction by pointing at any step, while the AI handles the detailed work of brainstorming strategies, decomposing the proof, discharging steps, and formalizing them in Lean.

To understand how different design choices affect proving performance, we conduct a systematic evaluation of the key components within \sysname{} across different proving skills and autoformalization routes. We carry out two complementary studies. The first is an evaluation on a FormalMATH subset comparing the NL-first and formalization-first skills, in which we evaluate Gemini~3.1~Pro, GPT-5.4, and Claude~Opus~4.7. The second is a focused study on two real PDE theorems, the L$^2$ error estimate for the semi-discrete upwind DG method and the Gauss--Radau projection reference-cell estimate, evaluated across three autoformalization routes, in which we test Gemini~3.1~Pro, GPT-5.4, Claude~Opus~4.7, and Claude~Sonnet~4 on their ability to produce fully verified Lean~4 proofs and to identify errors in deliberately incorrect proofs.

Our experiments reveal that current models can solve undergraduate-level problems at moderate success rates, and that writing a natural-language proof before formalization roughly doubles the strict pass rate. On the real PDE tasks, performance is highly sensitive to how the statement is formalized: given a well-formed Lean formalization, Claude~Opus~4.7 produces verified Lean~4 certificates on every attempt and GPT-5.4 follows closely, whereas the same models fail on version-incompatible formalizations. Performance also varies sharply across proving skills and error-detection capabilities. These findings suggest that while LLMs have made substantial progress on formal mathematics, bridging the gap to research-level mathematical reasoning remains an open challenge.

\section{Related Work}\label{sec:related}

\subsection{LLM-Based Theorem Proving}

The integration of AI-driven methods into formal proof has fostered a rich and rapidly evolving line of work. Early approaches primarily used language models for premise selection \citep{irving2016deepmath,wang2017premise}, while subsequent work explored direct proof generation via sequence-to-sequence models \citep{poluformal}. More recently, large language models (LLMs) have demonstrated unprecedented capabilities in mathematical reasoning. However, because verifying pure natural language proofs is inherently difficult, the mainstream paradigm has shifted toward integrating LLMs with formal languages, such as Lean \citep{moura2021lean} and Isabelle \citep{eberl2024isabelle}. Despite this shift, directly generating formal code remains a severe bottleneck for current LLMs due to the scarcity of formal training data. Consequently, recent advancements in this domain broadly bifurcate into two distinct categories.

The first category leverages natural language (informal mathematics) as an intermediate bridge. These approaches first generate a natural language proof and subsequently translate it into a formal language for verification. For instance, Draft-Sketch-Prove~\citep{jiangdraft} demonstrated that drafting an informal mathematical sketch to outline the high-level logical structure before attempting formalization significantly improves the model's success rate. Hilbert~\citep{varambally2025hilbert} further introduced a recursive refinement strategy that first attempts a direct formal proof and, upon verification failure, reverts to natural language to decompose the complex target into smaller, manageable sub-goals before recursively repeating this formalization-decomposition cycle. 

The second category confronts the generation challenge head-on by utilizing finetuning~\citep{wang2024theoremllama, wu2022autoformalization,hubert2025olympiad}, data synthesis~\citep{lin2025goedel}, and advanced inference strategy~\citep{thakur2023context,xin2025deepseek,chen2025seed} to directly enhance the model's capacity to produce formal tactics. For example, DeepSeek-Prover-V2 \citep{xin2025deepseek} utilizes Monte Carlo Tree Search (MCTS) guided by Lean's state feedback to explore complex proof trees. AlphaProof \citep{hubert2025olympiad} combined Gemini-based models with RL to solve competition-level geometry and algebra problems. Goedel-Prover-V2~\citep{lin2025goedel} further trains a prover capable of generation, correction, and expansion, rather than relying on pure single-pass generation.

\subsection{AI-Assisted Mathematical Research Systems}

Multi-agent reasoning systems \citep{du2023improving, wu2023mathchat,zhang2025debate4math} have demonstrated that collaborative debate among agents can improve reasoning quality, and interactive frameworks like ReAct \citep{yao2023react} enable iterative reasoning with tool use. Existing mathematical tool bases such as Mathlib~\citep{ulirsch2022blueprint} provide extensive community-maintained libraries, which are supported by rigorous verification environments like Lean. Building upon these environments, interactive proof systems such as LeanDojo \citep{yang2023leandojo} provides an open-source playground that enables fine-grained premise extraction, hard negative mining, and retrieval-augmented theorem proving, while Lean Copilot~\citep{song2025lean} assists humans by integrating LLM inference into Lean to suggest proof steps, complete sub-goals, and select premises.

Two systems are particularly relevant to our work. Numina-Lean-Agent \citep{liu2026numina} leverages the capabilities of Claude Code to autonomously orchestrate natural language proving and Lean verification. Rethlas \& Archon~\citep{ju2026automated} is a dual-agent system: Rethlas handles literature retrieval, proof planning, and heuristic refinement via natural language, while Archon compiles these informal plans into Lean and fills in the formal gaps. This framework successfully resolved an open Anderson conjecture in commutative algebra.

These systems share a common focus: given a statement, they aim to autonomously generate a complete formal proof. While the vision of fully automated, end-to-end theorem proving is ambitious and has driven significant progress, we argue that the ideal human-AI interaction paradigm should not seek to entirely replace human researchers. Instead, human experts are best suited for high-level directional design, leaving the tedious, low-level formalization tasks to AI. Consequently, our system, \sysname{}, heavily emphasizes active mathematician involvement. Within our system, humans supervise and continuously refine the overarching proof sketches, while delegating the fine-grained, step-level detail completion to AI. This division of labor ensures that the overall proving workflow remains highly controllable, mathematically intuitive, and ultimately more efficient.
\section{\sysname{}}\label{sec:system}

\begin{figure}[t]
    \centering
    \includegraphics[width=0.95\linewidth]{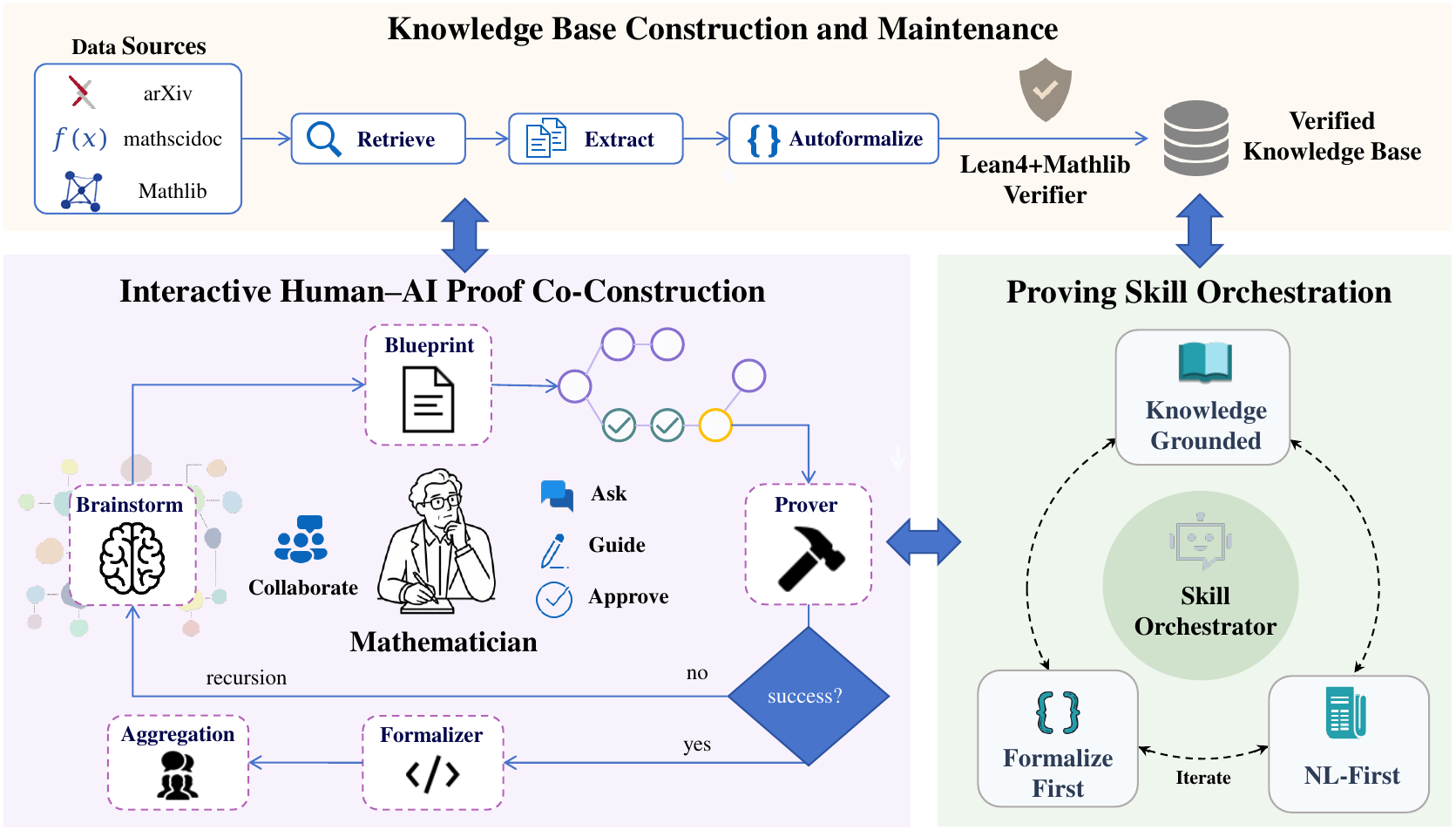}
    \caption{Architecture of \sysname{}. The mathematician interacts with three core capabilities, all connected to the Lean~4 verifier and a personal verified knowledge base.}
    \label{fig:architecture}
\end{figure}

\sysname{} is designed around a simple principle: the system should serve the mathematician, not the benchmark. This means supporting the full breadth of research activities within a single platform, with the mathematician retaining control over the process at every stage. We organize the system around three core capabilities: an interactive human--AI workbench that turns proof construction into a human-in-the-loop process centered on an editable proof blueprint; automated proving skill orchestration; and automated knowledge base construction and maintenance. Figure~\ref{fig:architecture} illustrates the overall architecture.

\subsection{Interactive Human--AI Proof Co-Construction}\label{sec:interactive}

The core capability of \sysname{} is an interactive workbench where a mathematician and AI agents co-construct a proof. Rather than delivering a single end-to-end proof, the system organizes the work as a human-in-the-loop pipeline of specialized agents centered on a shared, editable \emph{proof blueprint}. This keeps the mathematician in control of the high-level direction while the detailed sub-tasks are delegated to the agents. The workflow begins when the mathematician states a target proposition and proceeds through five agents.

\textbf{Brainstorm Agent.} Given the target, the \emph{Brainstorm Agent} proposes several candidate proof strategies, each a high-level natural-language sketch of a possible route to the result. The mathematician can edit these suggestions directly, merge or discard them, and ultimately selects one strategy to develop. This keeps the choice of overall approach, the step that most depends on mathematical taste, in human hands.

\textbf{Blueprint Agent.} From the selected strategy, the \emph{Blueprint Agent} generates a proof blueprint by decomposing the sketch into a graph of nodes, where each node is a self-contained sub-step, such as a lemma, a case, or an intermediate claim, and the edges encode logical dependencies. The blueprint is rendered as a flowchart annotated with each node's natural-language intent and current status (open, in progress, or verified), so that the mathematician can see at a glance which parts of the argument are settled and which remain open. Crucially, interaction is node-local and pointing-based: instead of locating a step through back-and-forth dialogue, the mathematician clicks directly into a node to inspect, edit, or act on that specific step. This makes large proofs navigable and gives the human and the agents a concrete shared workspace.

\textbf{Prover Agent.} The \emph{Prover Agent} is the discussion-time interface to the automated skill orchestration that will be introduced in Section~\ref{sec:proving}. For any node, the mathematician can designate the claim and ask the Prover Agent to discharge it fully automatically. When a step is too hard to close directly, the mathematician can instead recurse: the entire workflow is re-applied to that single node, asking the Brainstorm Agent for sub-strategies, generating a sub-blueprint, and proving the resulting finer sub-steps. This recursion can be repeated to arbitrary depth, letting a difficult proof be broken down until each leaf is automatically provable.

\textbf{Formalizer Agent.} The \emph{Formalizer Agent} maintains the correspondence between the informal blueprint and Lean~4. When the blueprint is first created, it emits a Lean sketch that mirrors the node structure, namely a skeleton of \texttt{have} statements whose unproven steps are filled with \texttt{sorry}. As each node is completed, it automatically translates that node's natural-language argument into Lean and validates it against the Lean~4 verifier, so that formal progress accumulates in lockstep with the informal proof.

\textbf{Aggregation Agent.} Once the blueprint is fully discharged, the \emph{Aggregation Agent} collects the per-node results and synthesizes them into a single coherent natural-language proof of the original target, suitable for inclusion in a paper. It linearizes the dependency graph, stitches the node-level arguments into connected prose, and pairs the exposition with the assembled Lean proof.

Throughout this pipeline, the system provides on-demand explanation: given any node or Lean fragment, it generates natural-language explanations at multiple granularities, from high-level strategy overviews to individual tactic justifications. A mathematician can, for example, ask ``why did you apply \texttt{induction} here?'' and receive both a formal account of the tactic's effect and an informal description of the underlying mathematical intuition.

\subsection{Proving Skill Orchestration}\label{sec:proving}

A central feature of \sysname{} is that proving is not a fixed sequential pipeline but rather an \emph{automated orchestration of skills} that adapt based on the current proof state, Lean feedback, and knowledge base content. The system maintains a pool of proving skills, such as NL-first (generate an informal sketch, then incrementally formalize), formalization-first (fully formalize the statement, then search for tactics via Lean's own proof state), and knowledge-grounded proving (retrieve analogous theorems and reusable lemmas from the user's knowledge base before and during proof construction).

Rather than executing these skills in a predetermined order, an orchestrator dynamically selects and composes skills at each step. The proving process typically proceeds as follows: the system begins by analyzing the formal statement and searching the knowledge base for relevant prior results. Based on this analysis, it selects an initial proving skill and generates a proof attempt. This attempt is then submitted to the Lean~4 environment, which returns either a success signal or detailed error information (e.g., type mismatches, missing hypotheses, incomplete tactics, unsolved goals). The orchestrator parses this feedback, diagnoses the failure mode, and decides on the next action, which may include retrying with a different skill, refining the current approach based on the specific error, searching for additional lemmas in the knowledge base, or requesting human intervention.

The key insight is that this is not a fixed workflow but a flexible, feedback-driven orchestration. In some cases, the NL-first skill may produce a clean proof on the first attempt. In others, the system may switch between multiple skills, incorporate knowledge base results mid-proof, and iteratively refine based on Lean feedback several times before arriving at a correct proof. The orchestrator learns from historical proof attempts which skills tend to succeed for which types of problems, improving its selection over time.

\subsection{Automated Knowledge Base Construction and Maintenance}

Given a research topic specified by the user, \sysname{} can retrieve relevant papers from multiple sources including Semantic Scholar, arXiv, and the Lean Mathlib, ranking them by relevance score, citation count, and formalization potential. For each retrieved paper, an extraction agent identifies formal definitions and their dependencies, theorem statements including all hypotheses and conclusions, proof sketches and lemma chains, and the notation conventions used by the authors. The extracted content is then translated into Lean~4 code using LLM prompting with access to the Mathlib library as context, following a dependency-aware ordering where definitions are formalized first, then auxiliary lemmas, then main theorems. Each formalized Lean file is compiled and verified against the Lean~4 type checker, and only content that passes compilation enters the knowledge base.

Crucially, each knowledge base entry stores both the verified Lean~4 code and the natural language proof process that led to it, including the informal reasoning chain, key insights at each step, and the correspondence between natural language arguments and formal tactics. This dual representation serves two purposes: it enables the system to provide natural language explanations when a mathematician retrieves a lemma, and it allows the agents in Section~\ref{sec:interactive} to reason about proofs in both informal and formal terms.

Verified entries are indexed by topic tags, theorem names, dependency graphs, and natural language summaries to support flexible retrieval. The knowledge base is personal: it grows organically as the mathematician processes papers and proves theorems, accumulating a reusable repository tailored to their research interests. The system also supports incremental maintenance: when the underlying Lean environment (e.g., Mathlib) is updated, existing entries are automatically re-verified, and any incompatibilities are flagged for the user.
\section{FormalMATH Benchmark Experiments}\label{sec:exp}\label{sec:formalmath-exp}

We first evaluate \sysname{} in a controlled benchmark setting using a sampled subset of FormalMATH. This section isolates the effect of proving skill choice on undergraduate-level formal mathematics, and then studies whether the full interactive proof-blueprint workflow can recover hard cases unsolved by GPT-5.4 through recursive refinement of unresolved local proof obligations.

\subsection{FormalMATH Experimental Setup}

\textbf{Models and routes.} The FormalMATH study evaluates Gemini~3.1~Pro, GPT-5.4, and Claude~Opus~4.7. Each model is tested under two proving skills from Section~\ref{sec:proving}: formalization-first (FF), where the model works from the formal statement directly, and NL-first (NL), where it first produces a natural-language proof sketch and then translates the argument into Lean.

\textbf{FormalMATH subset.} We sample 21 problems from FormalMATH to compare the proving skills, with exactly three problems from each of seven domains: Algebra, Differentiation, Integral, Multivariable Calculus, Other, Precalculus, and Sequences and Series. Each problem contains a natural-language statement, a formalized Lean statement, and a reference solution. The sampled problems are listed in Appendix~\ref{app:formalbench-samples}. This gives $21 \times 3 \times 2 = 126$ proof attempts.

\textbf{GPT-unsolved subset.}
To evaluate the full our interactive system beyond a single baseline prompting route, we select 12 cases that GPT-5.4 failed to solve:
\texttt{u-math\_862}, \texttt{u-math\_645}, \texttt{u-math\_166}, \texttt{u-math\_839}, \texttt{u-math\_93}, \texttt{u-math\_237}, \texttt{hardmath\_162}, \texttt{DEMIMathAnalysis\_86}, \texttt{DEMIMathAnalysis\_26}, \texttt{DEMIMathAnalysis\_37}, \texttt{DEMIMathAnalysis\_19}, and \texttt{DEMIMathAnalysis\_75}. In this setting the system is allowed to use the interactive proof-blueprint workflow: when a first formalization fails, the unresolved proof obligations are exposed as local nodes, searched against standard mathematical facts, and recursively formalized until either Lean verifies the node or the node remains open.

\textbf{Verification protocol.}
We verify FormalMATH proofs with Lean~4.30.0-rc2 and Mathlib~v4.30.0-rc2 under a 1200-second timeout. We report a \emph{strict} pass rate: a proof passes only if \texttt{lake env lean} accepts the file, the final proof is nonempty, and the file contains no \texttt{sorry}/placeholder proof. This strict rule matters: Claude~Opus~4.7 has 6/21 raw verifier passes under Formalization-First, but three are empty formal-proof outputs and are therefore counted as failures.

\subsection{FormalMATH Results}

Table~\ref{tab:formalbench-overall} shows the strict pass rates. The main pattern is simple: generating a natural-language proof before Lean translation substantially improves reliability. The NL skill reaches 20/63 overall, exactly double FF's 10/63. The improvement is consistent across all three models, with gains of +14.29\% for Gemini and GPT and +19.05\% for Opus after the empty-proof correction.

\begin{table}[t]
\centering
\caption{Strict pass rates on the 21-problem FormalMATH subset.}
\label{tab:formalbench-overall}
\small
\begin{tabular}{lccc}
\toprule
\textbf{Model} & \textbf{FF} & \textbf{NL} & \textbf{Gain} \\
\midrule
Gemini~3.1~Pro & 4/21 (19.05\%) & 7/21 (33.33\%) & +14.29 \\
GPT-5.4 & 3/21 (14.29\%) & 6/21 (28.57\%) & +14.29 \\
Claude~Opus~4.7 & 3/21 (14.29\%) & 7/21 (33.33\%) & +19.05 \\
\midrule
\textbf{Overall} & \textbf{10/63 (15.87\%)} & \textbf{20/63 (31.75\%)} & \textbf{+15.87} \\
\bottomrule
\end{tabular}
\end{table}

The domain-level heatmap in Table~\ref{tab:formalbench-heatmap} gives a more informative view. The NL skill is not uniformly better because the benchmark mixes very different formalization burdens. It gives large gains in Differentiation, where FF solves none of the nine attempts but NL solves six, and in Precalculus, where the per-skill score increases from 5/9 to 7/9. In contrast, Integral, Other, and Sequences and Series remain at 0/18 across both skills and all models. These zero rows are important: they show that failures are not merely model-choice noise, but often reflect missing formal infrastructure, brittle theorem statements, or proof obligations that current single-turn prompting does not bridge.

\begin{table}[t]
\centering
\caption{Domain heatmap for FormalMATH. Darker green indicates higher pass rate.}
\label{tab:formalbench-heatmap}
\small
\setlength{\tabcolsep}{5pt}
\begin{tabular}{lcccccccc}
\toprule
\textbf{Domain} &
\makecell{\textbf{FF}\\\textbf{Total}} &
\makecell{\textbf{NL}\\\textbf{Total}} &
\makecell{\textbf{Gemini}\\\textbf{FF}} &
\makecell{\textbf{Gemini}\\\textbf{NL}} &
\makecell{\textbf{GPT}\\\textbf{FF}} &
\makecell{\textbf{GPT}\\\textbf{NL}} &
\makecell{\textbf{Opus}\\\textbf{FF}} &
\makecell{\textbf{Opus}\\\textbf{NL}} \\
\midrule
Integral & \heatZero{0/9} & \heatZero{0/9} & \heatZero{0/3} & \heatZero{0/3} & \heatZero{0/3} & \heatZero{0/3} & \heatZero{0/3} & \heatZero{0/3} \\
Other & \heatZero{0/9} & \heatZero{0/9} & \heatZero{0/3} & \heatZero{0/3} & \heatZero{0/3} & \heatZero{0/3} & \heatZero{0/3} & \heatZero{0/3} \\
Seq./Series & \heatZero{0/9} & \heatZero{0/9} & \heatZero{0/3} & \heatZero{0/3} & \heatZero{0/3} & \heatZero{0/3} & \heatZero{0/3} & \heatZero{0/3} \\
Multivariable Calc. & \heatLow{2/9} & \heatLow{3/9} & \heatLow{1/3} & \heatLow{1/3} & \heatLow{1/3} & \heatLow{1/3} & \heatZero{0/3} & \heatLow{1/3} \\
Differentiation & \heatZero{0/9} & \heatHigh{6/9} & \heatZero{0/3} & \heatHigh{2/3} & \heatZero{0/3} & \heatHigh{2/3} & \heatZero{0/3} & \heatHigh{2/3} \\
Algebra & \heatLow{3/9} & \heatMid{4/9} & \heatLow{1/3} & \heatLow{1/3} & \heatLow{1/3} & \heatLow{1/3} & \heatLow{1/3} & \heatHigh{2/3} \\
Precalculus & \heatMid{5/9} & \heatHigh{7/9} & \heatHigh{2/3} & \heatFull{3/3} & \heatLow{1/3} & \heatHigh{2/3} & \heatHigh{2/3} & \heatHigh{2/3} \\
\bottomrule
\end{tabular}
\end{table}

Failure analysis further supports this interpretation. The most common blocking category is residual \texttt{sorry}: 19 times under FF and 23 times under NL. Thus, the models often understand enough to state a plausible proof plan but cannot discharge the Lean proof obligations. The NL skill reduces this problem for tractable domains by giving the Lean stage a structured proof skeleton, but it does not solve domains whose formal statements require substantial libraries or nontrivial analytic machinery.
Appendix~\ref{app:hardmath734} gives a concrete NL-skill example from the Integral domain: all three models identify the correct endpoint Laplace-method argument for \texttt{hardmath\_734}, yet each final Lean file still contains \texttt{sorry}. This case illustrates why strict verification can disagree with a superficially convincing natural-language derivation.

\subsection{Recovering Failures with \sysname{}}

Table~\ref{tab:mathcopilot-formalmath-followup} reports the recovery experiment on the 12 difficult FormalMATH cases unsolved by GPT-5.4. In the first system pass, nine of the twelve cases are fully formalized and accepted by Lean under the same strict rule used above. Three cases remain open: \texttt{u-math\_839}, \texttt{u-math\_237}, and \texttt{hardmath\_162}. The failure mode of \texttt{hardmath\_162} is qualitatively different from the other two: the system produces a complete natural-language proof of the endpoint Laplace asymptotic, but the Lean attempt stalls at local analytic and numerical sublemmas. When those sublemmas are recursively exposed as proof-blueprint nodes, the system eventually formalizes the endpoint Laplace statement completely, yielding a verified Lean certificate in \texttt{tex/lean/Main.lean}. Counting this lemma-refinement stage, \sysname{} verifies 10/12 of the targeted GPT-unsolved cases.

\begin{table}[t]
\centering
\caption{\sysname{}'s performance on twelve GPT-unsolved FormalMATH cases. ``First pass'' counts the initial full-system attempt; ``after refinement'' allows recursive proof-blueprint formalization of unsolved local lemmas.}
\label{tab:mathcopilot-formalmath-followup}
\scriptsize
\setlength{\tabcolsep}{3pt}
\begin{tabular}{p{0.24\textwidth}p{0.18\textwidth}p{0.18\textwidth}p{0.23\textwidth}}
\toprule
\textbf{Problem} & \textbf{Domain} & \textbf{First pass} & \textbf{After refinement} \\
\midrule
\texttt{u-math\_862} & Precalculus & \passmark & \passmark \\
\texttt{u-math\_645} & Algebra & \passmark & \passmark \\
\texttt{u-math\_166} & Integral & \passmark & \passmark \\
\texttt{u-math\_839} & Multivariable Calc. & \failmark & \failmark \\
\texttt{u-math\_93} & Multivariable Calc. & \passmark & \passmark \\
\texttt{u-math\_237} & Sequences/Series & \failmark & \failmark \\
\texttt{hardmath\_162} & Integral & \failmark & \passmark \\
\texttt{DEMIMathAnalysis\_86} & Other & \passmark & \passmark \\
\texttt{DEMIMathAnalysis\_26} & Other & \passmark & \passmark \\
\texttt{DEMIMathAnalysis\_37} & Differentiation & \passmark & \passmark \\
\texttt{DEMIMathAnalysis\_19} & Sequences/Series & \passmark & \passmark \\
\texttt{DEMIMathAnalysis\_75} & Other & \passmark & \passmark \\
\midrule
\textbf{Total} & -- & \textbf{9/12 (75.0\%)} & \textbf{10/12 (83.3\%)} \\
\bottomrule
\end{tabular}
\end{table}

Table~\ref{tab:hardmath162-system-comparison} gives a closer view of the most informative recovery case, \texttt{hardmath\_162}. In the baseline FormalMATH experiment, all six Gemini/GPT/Opus solutions on this problem, across both FF and NL settings, end with the same failure category: the generated Lean file contains a top-level \texttt{sorry}.The natural-language behavior is more nuanced. Some attempts identify the endpoint-Laplace route, while others correctly notice that the autoformalized target is semantically brittle: the Lean statement uses an outer parameter \(x\), then shadows it inside the function whose limit is being taken, and asks for convergence to a rounded expression depending on the outer \(x\). In either case, the baseline attempts do not convert the Laplace argument into checkable local proof obligations.

\begin{table}[t]
\centering
\caption{Detailed comparison on \texttt{hardmath\_162}. Baseline evidence comes from the prior FormalMATH runs in \texttt{formalbench\_project}.}
\label{tab:hardmath162-system-comparison}
\scriptsize
\setlength{\tabcolsep}{3pt}
\begin{tabular}{p{0.15\textwidth}p{0.23\textwidth}p{0.25\textwidth}p{0.24\textwidth}}
\toprule
\textbf{Attempt} & \textbf{Mathematical behavior} & \textbf{Lean behavior} & \textbf{Main limitation or strength} \\
\midrule
Gemini/GPT/Opus baselines, FF and NL &
Often identify the endpoint-Laplace structure or flag the rounded target as problematic. &
All six generated Lean files instantiate a direct \texttt{Tendsto} target for the rounded expression and end with \texttt{sorry}; verifier category: \texttt{contains\_sorry}. &
No decomposition of derivative bounds, endpoint separation, integration-by-parts remainder, or numerical enclosures into verified subgoals. \\
\sysname{} first pass &
Finds the correct left-endpoint Laplace proof strategy and produces a coherent natural-language derivation. &
Fails strict verification because local analytic and numerical lemmas remain unresolved. &
Failure is localized to explicit proof-blueprint nodes rather than hidden in one monolithic placeholder. \\
\sysname{} refinement &
Recursively formalizes the unresolved endpoint-Laplace lemma, including exact asymptotics and rounded-constant bounds. &
Produces a complete Lean certificate with no placeholder proof in the endpoint block. &
Turns the informal appeal to Laplace's method into machine-checkable monotonicity, integration-by-parts, remainder, and interval-arithmetic obligations. \\
\bottomrule
\end{tabular}
\end{table}

The \texttt{hardmath\_162} refinement is particularly informative. The unsolved obligation is not a high-level proof strategy failure but the formalization of an endpoint Laplace lemma:
\[
I(x)=\int_{-0.7}^{0.5}(-0.9t^2)e^{x\phi(t)}\,dt,
\qquad
\phi(t)=1.7t^4-0.3t-0.6\cos t-2\arctan t+2.6,
\]
with the target asymptotic
\[
I(x)\sim
\frac{(-0.9)(-0.7)^2 e^{x\phi(-0.7)}}{x(-\phi'(-0.7))}
\qquad (x\to+\infty),
\]
together with the numerical bounds
\[
|\phi(-0.7)-3.98|\le0.001,\qquad
\left|\frac{(-0.9)(-0.7)^2}{-\phi'(-0.7)}+0.10\right|\le0.002.
\]
These are standard consequences of the endpoint form of Laplace's method plus concrete interval estimates. The system decomposes them into derivative identities, endpoint separation, integration-by-parts, remainder bounds, and numerical inequalities, after which Lean verifies the final theorem. Appendix~\ref{app:hardmath162-refinement} gives the detailed evidence.

The refined statement is also mathematically sharper than the displayed benchmark approximation. The proof does not assert the strict asymptotic equivalence
\[
I(x)\sim -\frac{0.10 e^{3.98x}}{x},
\]
which would be sensitive to the rounded exponent. Instead, it proves the exact endpoint formula
\[
I(x)\sim
\frac{-0.9(-0.7)^2}{-\phi'(-0.7)}
\frac{e^{x\phi(-0.7)}}{x},
\]
and separately proves that \(\phi(-0.7)\) is within \(0.001\) of \(3.98\) and that the leading coefficient is within \(0.002\) of \(-0.10\). This separation between exact asymptotics and rounded numerical presentation is precisely the kind of distinction that the proof-blueprint workflow makes explicit.

\section{PDE Theorem Proving Experiments}\label{sec:real-pde}

We next evaluate \sysname{} on two real theorems from numerical PDE analysis. This setting is deliberately more demanding than the FormalMATH benchmark: the statements come from actual domain-specific analysis, the proofs rely on numerical-method concepts that are not fully developed in Mathlib, and success depends on choosing a suitable formalization boundary rather than merely finding a short tactic proof. We therefore treat the PDE study as a separate real-data evaluation of whether current models can help with research-level mathematical formalization.

\subsection{PDE Experimental Setup}

\textbf{Theorems.} We use two real theorems from numerical PDE analysis. They are drawn from the standard discontinuous Galerkin (DG) analysis pipeline for transport-dominated PDEs. Linear advection is the canonical scalar model for wave propagation and conservation-law transport; upwind DG schemes trace back to finite-element discretizations of neutron-transport equations and later became a central family of high-order methods for hyperbolic conservation laws~\citep{lesaint1974finite,cockburn1989tvb,cockburn2001runge}. The first is the semi-discrete upwind DG error estimate for linear advection,
\begin{equation}
  \label{eq:dg-advection-error}
  \|u(\cdot,T)-u_h(\cdot,T)\|_{L^2(0,1)}
  \le C h^{K+1},
\end{equation}
proved by splitting \(u-u_h=(u-\Pi_hu)+(\Pi_hu-u_h)\), using Galerkin orthogonality, an energy estimate, and Gronwall's inequality, following the classical error-analysis template for scalar hyperbolic DG methods~\citep{johnson1986analysis,zhang2004error}. The second is the Gauss--Radau reference-cell projection estimate,
\begin{align}
  \label{eq:gauss-radau-reference}
  \|\hat \Pi \hat w-\hat w\|_{L^2(\hat I)}
  &\le C |\hat w|_{H^{k+1}(\hat I)}, \\
  \label{eq:gauss-radau-physical}
  \|P w-w\|_{L^2(I_i)}
  &\le C h_i^{k+1}|w|_{H^{k+1}(I_i)}.
\end{align}
which represents the local approximation ingredient used in those DG analyses: endpoint matching is aligned with the one-sided upwind flux, while the \(L^2\) bounds are consequences of polynomial approximation, Bramble--Hilbert type estimates, and affine scaling from a reference cell to a physical mesh cell~\citep{bramble1970estimation,ciarlet2002finite,zhang2004error}. Appendix~\ref{app:pde-cases} states the two PDE cases in more detail.

\textbf{Routes and models.} For these two theorems, we adopt three routes. In the \emph{GPT-Formalized} and \emph{Opus-Formalized} routes, GPT-5.4 and Claude~Opus~4.7 respectively formalize the statement into a Lean stub, which is then handed to each model to prove. In the \emph{Direct} route, each model formalizes the statement and writes both the natural-language and Lean proofs on its own. We evaluate Gemini~3.1~Pro, GPT-5.4, and Claude~Opus~4.7 on all three routes, and Claude~Sonnet~4 on the GPT-Formalized route. This yields 20 Lean proof attempts.

\textbf{Verification protocol and scope.} We use Lean~4.30.0 and Mathlib~v4.30.0 under a 600-second timeout. Current Mathlib does not contain a full formal development of DG methods, Sobolev estimates, or Gauss--Radau projection theory. Therefore, successful PDE files often encode the deep analytic ingredients as \texttt{opaque} declarations or \texttt{axiom}s and then verify the formal assembly step in Lean. We count these as successful Lean certificates, not as end-to-end formalizations of PDE analysis from first principles. Consequently, the PDE pass rates in Table~\ref{tab:pde} are not directly comparable to the end-to-end strict pass rates in Table~\ref{tab:formalbench-overall}.

\textbf{Wrong-proof diagnostic.} Because proof review is an important part of real mathematical work, we also test wrong-proof detection by giving Gemini~3.1~Pro and GPT-5.4 deliberately flawed proofs of the two PDE theorems. Appendices~\ref{app:pde-proof-showcase} and~\ref{app:pde-wrong-proofs} summarize the proof-output patterns and wrong-proof diagnostics.

\subsection{Proof Synthesis Results}

Table~\ref{tab:pde} summarizes the PDE proof synthesis results. The GPT-Formalized route performs poorly not because the theorems are mathematically impossible, but because the GPT-produced formalizations are associated with Lean-version and declaration-boundary incompatibilities, including generated declarations that should instead be represented as explicit \texttt{opaque} or \texttt{axiom} assumptions in the tested Lean~4.29 setting. This formalization/version issue causes Gemini, GPT-5.4, and Sonnet to fail both theorems. Claude~Opus~4.7 is the only model that adapts the formalization and produces assembly certificates for both. The Opus-Formalized route yields assembly certificates for all six tested attempts. The Direct route is intermediate: GPT-5.4 and Claude~Opus~4.7 produce assembly certificates for both theorems under their own declared assumptions, while Gemini~3.1~Pro produces Lean files with logical/tactic-level failures.

\begin{table}[t]
\centering
\caption{PDE theorem proving results. Each entry is Lean-checked certificates out of two. ``Syntax'' denotes Lean-version/formalization incompatibility; ``Logic'' denotes a syntactically valid proof attempt that fails during proof checking. These are not end-to-end formalizations of PDE analysis from first principles.}
\label{tab:pde}
\small
\begin{tabular}{lcccc}
\toprule
\textbf{Model} & \makecell{\textbf{GPT}\\\textbf{Formalized}} & \makecell{\textbf{Opus}\\\textbf{Formalized}} & \textbf{Direct} & \textbf{Total} \\
\midrule
Gemini~3.1~Pro & 0/2 (Syntax) & 2/2 (assembly) & 0/2 (Logic) & 2/6 \\
GPT-5.4 & 0/2 (Syntax) & 2/2 (assembly) & 2/2 (assembly) & 4/6 \\
Claude~Opus~4.7 & 2/2 (assembly) & 2/2 (assembly) & 2/2 (assembly) & 6/6 \\
Claude~Sonnet~4 & 0/2 (Syntax) & -- & -- & 0/2 \\
\midrule
\textbf{Per-route total} & \textbf{2/8 (25\%)} & \textbf{6/6 (100\%)} & \textbf{4/6 (66.67\%)} & \textbf{12/20 (60\%)} \\
\bottomrule
\end{tabular}
\end{table}

These results reveal a different failure mode from the FormalMATH experiment. In the PDE setting, the strongest closed models can produce mathematically coherent natural-language proofs: they mention the correct DG error decomposition, the admissibility of the discrete error $\xi=\Pi_hu-u_h$ as a test function, energy stability, Gronwall's inequality, and the Bramble--Hilbert/scaling argument for Gauss--Radau projection. The main bottleneck is turning that understanding into a version-correct and appropriately abstract Lean. Successful proofs typically separate the work into two layers: analytic PDE facts are declared as explicit assumptions, while Lean verifies the algebraic combination, norm inequality, or scaling step.
Appendix~\ref{app:pde-proof-showcase} summarizes representative PDE proof-output patterns and identifies which analytic ingredients are treated as \texttt{opaque} declarations or \texttt{axiom}s.

We also include an out-of-distribution diagnostic with two open-source theorem-proving models, DeepSeek-Prover-V2~\citep{ren2025deepseek} and Goedel-Prover-V2~\citep{lin2025goedel}, on the DG theorem. Both fail to produce valid Lean~4 code (0/2). Their outputs remain in natural-language meta-analysis and then repeat until truncation. This suggests that specialization on competition-style formal mathematics does not automatically transfer to PDE/numerical-analysis reasoning.

\subsection{Wrong-Proof Detection}

We additionally evaluate whether models can act as mathematical proof reviewers rather than proof constructors. For each PDE theorem, we provide a polished but deliberately flawed proof and ask Gemini~3.1~Pro and GPT-5.4 to decide whether the argument is correct and to locate the failure.

\begin{table}[t]
\centering
\caption{Wrong-proof detection on the two PDE examples. ``Judgment'' records whether the model correctly rejects the flawed proof; ``coverage'' summarizes which planted issue is explicitly identified.}
\label{tab:wrong-proof-detection}
\small
\begin{tabular}{p{0.18\textwidth}p{0.32\textwidth}p{0.32\textwidth}c}
\toprule
\textbf{Model} & \textbf{DG $L^2$ estimate} & \textbf{Gauss--Radau estimate} & \textbf{Judgment} \\
\midrule
Gemini~3.1~Pro &
Rejects the proof and identifies the decisive illegal test function $e=u-u_h\notin U_h^K$; it also notes that interface terms form a jump contribution rather than canceling. &
Rejects the proof and identifies that $F(\hat w)=\|\hat\Pi\hat w-\hat w\|_{L^2}$ is a norm of an error, not a linear functional. &
\passmark \\
GPT-5.4 &
Rejects the proof, explicitly states that Galerkin orthogonality holds only for $\varphi_h\in U_h^K$, and identifies the illegal choice $\varphi=e$. &
Rejects the proof, explains the nonlinearity of $F(\hat w)$, and proposes a corrected operator or seminorm-based Bramble--Hilbert argument. &
\passmark \\
\bottomrule
\end{tabular}
\end{table}

Both models correctly reject all four model-theorem pairs. The diagnostic cases in Appendix~\ref{app:pde-wrong-proofs} show that critique can be easier than construction: even when Lean proof generation depends on abstracting away PDE theory with \texttt{opaque} declarations or \texttt{axiom}s, the models can still recognize invalid mathematical moves in natural-language proofs. This supports using \sysname{}'s Formalizer Agent as a complementary mode to proof synthesis, especially for domain-specific arguments whose full formal library support is still incomplete.

\section{Discussion}\label{sec:discussion}

Our results highlight four key takeaways. Current models show meaningful capability on undergraduate-level formal mathematics, but this capability is highly sensitive to the autoformalization pipeline and to the domain's available formal infrastructure. The NL-first skill consistently outperforms formalization-first on the FormalMATH subset, validating the intuition that informal reasoning provides a valuable scaffold for formalization. The recovery study further shows that an interactive proof-blueprint loop can recover many hard cases missed by a strong baseline: among twelve GPT-5.4-missed cases, the system produces strict Lean certificates for 9/12 immediately and 10/12 after local lemma refinement. Finally, the gap between benchmark proofs and PDE certificates, together with the wrong-proof diagnostic, underscores that proof construction, formal certification, and mathematical critique remain distinct skills.

These findings carry practical implications: investing in better autoformalization pipelines may yield larger returns than scaling up model size, and evaluation of formal mathematics systems should include domain-specific problems alongside curated benchmarks.

\section{Conclusion and Future Work}\label{sec:conclusion}

We presented \sysname{}, a human-in-the-loop system embodying a human--AI symbiotic paradigm for mathematical research, which unifies knowledge construction, automated proving skill orchestration, and interactive human--AI collaboration. Our experiments across four state-of-the-art LLMs reveal that these models demonstrate strong capability on undergraduate-level formal mathematics, that the NL-first proving skill achieves higher accuracy than formalization-first, and that autoformalization quality has a decisive impact on proving success. At the same time, performance on real PDE theorems and error detection tasks shows that substantial challenges remain for domain-specific mathematical reasoning.

Several directions for future work are promising. Integrating with community resources such as Mathlib to bootstrap larger knowledge bases would enable cross-community knowledge sharing and reduce the cold-start problem for new users. Training the skill orchestrator to learn from historical proof attempts could develop domain-specific skill preferences that improve with use. Supporting collaborative knowledge bases across research groups, with access control, versioning, and merge capabilities, would extend the system from individual to team-level use. Extending beyond Lean~4 to support Coq, Isabelle, and other proof assistants would enable cross-system knowledge transfer. Finally, conducting rigorous user studies with larger participant pools and controlled experiments would quantify the impact on research productivity.

\bibliographystyle{plainnat}
\bibliography{refs}

\clearpage
\appendix

\section{Appendix}\label{app:case-studies}

This appendix provides compact case-level details supporting the aggregate results. We first present seven representative FormalMATH problems, one from each sampled domain, to make the benchmark composition concrete. We then analyze the \texttt{hardmath\_734} Integral case as a representative NL-skill failure and the \texttt{hardmath\_162} refinement as a representative \sysname{} recovery case, before turning to the two PDE diagnostics used for proof synthesis and wrong-proof detection.

\subsection{Seven Representative FormalMATH Problems}\label{app:formalbench-samples}

To make the benchmark composition transparent, we present one representative problem from each sampled FormalMATH domain. These examples expose the heterogeneous nature of the subset, which ranges from algebraic simplification and elementary differentiation to asymptotic integration, multivariable integration, functional-equation nonexistence, and sequence-space convergence.

\begin{formalbenchproblem}{Problem 1: Algebra}{u-math\_551}
\problemstatement
Subtract the rational expression, then simplify:
\[
  \frac{12}{2\cdot q}-\frac{6}{3\cdot p}.
\]

\problemtarget
Prove that the simplified expression is
\[
  \frac{6\cdot p-2\cdot q}{p\cdot q}.
\]
\end{formalbenchproblem}

\begin{formalbenchproblem}{Problem 2: Differentiation}{DEMIMathAnalysis\_37}
\problemstatement
Suppose that \(f\in C^2(0,\infty)\) converges to \(\alpha\) as \(x\to\infty\), and that
\[
  f''(x)+\lambda f'(x)
\]
is bounded above for some constant \(\lambda\).

\problemtarget
Show that
\[
  f'(x)\to 0
  \qquad (x\to\infty).
\]
\end{formalbenchproblem}

\begin{formalbenchproblem}{Problem 3: Integral}{hardmath\_162}
\problemstatement
Consider the integral
\[
  I(x)=
  \int_{-0.7}^{0.5}
  (-0.9t^{2})
  e^{-x\left(-1.7t^{4}+0.3t+0.6\cos(t)+2.0\operatorname{atan}(t)-2.6\right)}
  \,dt.
\]
Develop an analytical formula for \(I(x)\) that is accurate as \(x\to\infty\).

\problemtarget
Prove the leading asymptotic form
\[
  I(x)\approx -\frac{0.10 e^{3.98x}}{x}.
\]
\end{formalbenchproblem}

\begin{formalbenchproblem}{Problem 4: Multivariable Calculus}{u-math\_839}
\problemstatement
Evaluate the triple integral
\[
  \iiint_E z\,dV,
\]
where \(E\) is the region
\[
  E=
  \left\{
  (x,y,z)\ \middle|\ 
  \begin{aligned}
    -y &\le x \le y,\\
    0 &\le y \le 1,\\
    0 &\le z \le 1-x^4-y^4
  \end{aligned}
  \right\}.
\]

\problemtarget
Prove that
\[
  \iiint_E z\,dV=\frac{113}{450}.
\]
\end{formalbenchproblem}

\begin{formalbenchproblem}{Problem 5: Other}{DEMIMathAnalysis\_26}
\problemstatement
Consider continuous functions
\[
  f,g,h:\mathbb{R}\to\mathbb{R}.
\]
The claimed functional equation is
\[
  h(f(x)+g(y))=xy
\]
for all \((x,y)\in\mathbb{R}^2\).

\problemtarget
Show that no such continuous functions \(f\), \(g\), and \(h\) exist.
\end{formalbenchproblem}

\begin{formalbenchproblem}{Problem 6: Precalculus}{u-math\_862}
\problemstatement
Calculate the derivative
\[
  \frac{d}{dx}\left(\log_x(a)\right)
\]
under the assumptions
\[
  x>0,\qquad a>0,\qquad x\ne 1,\qquad a\ne 1.
\]

\problemtarget
Prove that
\[
  \frac{d}{dx}\left(\log_x(a)\right)
  =
  -\frac{\ln(a)}{x(\ln x)^2}.
\]
\end{formalbenchproblem}

\begin{formalbenchproblem}{Problem 7: Sequences and Series}{DEMIMathAnalysis\_19}
\problemstatement
Suppose that
\[
  \sum_{n=1}^{\infty} a_n b_n
\]
converges for every sequence \(\{b_n\}\) satisfying
\[
  \sum_{n=1}^{\infty} b_n^2<\infty.
\]

\problemtarget
Prove that
\[
  \sum_{n=1}^{\infty} a_n^2
\]
also converges.
\end{formalbenchproblem}

\subsection{Case C: \texorpdfstring{\texttt{hardmath\_734}}{hardmath_734} NL-First Outputs}\label{app:hardmath734}

\begin{hardmathbluebox}{Problem: \texttt{hardmath\_734}\hfill Domain: Integral}
Consider the integral
\begin{equation}
I(x)=\int_{-0.8}^{-0.1}
\bigl(2.7t^5+1.5\sin(t)-1.1\cos(t)+0.5\bigr)
 e^{+x\left(-0.5t^5+4.1\cos(t)+1.9\operatorname{atan}(t)\right)}\,dt.
\end{equation}
Develop an analytical formula for \(I(x)\) that is accurate as \(x\to\infty\). Prove that the answer is:
\[
  \boxed{I(x)\approx -\frac{0.32e^{3.89x}}{x}}.
\]
\end{hardmathbluebox}

\begin{hardmathbluebox}{REFERENCE SOLUTION}
The integral is of the form
\begin{equation}
  I(x)=\int_a^b g(t)e^{+xf(t)}\,dt,
\end{equation}
where \(a=-0.8\), \(b=-0.1\),
\[
  g(t)=2.7t^5+1.5\sin(t)-1.1\cos(t)+0.5,
  \qquad
  f(t)=-0.5t^5+4.1\cos(t)+1.9\operatorname{atan}(t).
\]
This means we can use Laplace's method to develop an analytical approximation in the limit that \(x\to\infty\).
In this limit, the integral will be dominated by the integrand near the maximum of \(f(t)\) within the bounds \([-0.8,-0.1]\). So, to simplify the integral, we will expand the integrand around this maximum.

However, it is impossible to compute the maximum of \(f(t)\) within these bounds analytically. We, therefore, find the maximum numerically with the dual annealing algorithm. Using a few iterations of dual annealing, we find that the maximum of \(f(t)\) occurs at \(t_0=[-0.1]\).

Since the integral is dominated by the value of the integrand near \(t_0=b\), we Taylor expand the integrand around this point:
\begin{equation}
I(x)=\int_a^b
\bigl(g(b)+(t-b)g'(b)+\cdots\bigr)
 e^{+x\left(f(b)+(t-b)f'(b)+\cdots\right)}\,dt.
\end{equation}
We can then approximate
\begin{equation}
I(x)\approx\int_a^b g(b)e^{+x\left(f(b)+(t-b)f'(b)\right)}\,dt.
\end{equation}
Pulling out the constant in the integral gives
\begin{equation}
I(x)\approx g(b)e^{+xf(b)}\int_a^b e^{+x(t-b)f'(b)}\,dt.
\end{equation}
We perform the change of variables \(u=x(t-b)|f'(b)|\), rewriting the integral as
\begin{equation}
I(x)\approx g(b)e^{+xf(b)}
\int_{x(a-b)|f'(b)|}^{0}\frac{1}{x|f'(b)|}e^u\,du.
\end{equation}
Since we are interested in a solution for \(I(x)\) as \(x\to\infty\), we can simplify the integral by approximating the lower bound as
\begin{equation}
I(x)\approx g(b)e^{+xf(b)}\frac{1}{x|f'(b)|}\int_{-\infty}^{0}e^u\,du.
\end{equation}
Solving the integral and evaluating, we find that
\begin{equation}
  \boxed{I(x)\approx -\frac{0.32e^{3.89x}}{x}}.
\end{equation}
\end{hardmathbluebox}


\begin{hardmathredbox}{NL-First Summary}
All three NL-First outputs identify the correct high-level asymptotic method: endpoint Laplace method near \(b=-0.1\). However, all three Lean~4 files end with \texttt{sorry}. Under the strict evaluation rule, any proof containing \texttt{sorry} is marked as failed, even when Lean reports only a warning.
\end{hardmathredbox}


\hardmathmodelheading{Gemini 3.1 Pro NL-First Natural-Language Proof}

\hardmathsmalllabel{Given:}
The integral
\[
I(x)=\int_{-0.8}^{-0.1}
\bigl(2.7t^5+1.5\sin(t)-1.1\cos(t)+0.5\bigr)
 e^{x\left(-0.5t^5+4.1\cos(t)+1.9\operatorname{atan}(t)\right)}\,dt.
\]
We are tasked with developing an analytical formula for \(I(x)\) that is accurate as \(x\to\infty\) and proving that
\[
I(x)\approx -\frac{0.32e^{3.89x}}{x}.
\]

\hardmathsmalllabel{Derivation:}
Let us define the amplitude function \(f(t)\) and the phase function \(g(t)\) from the integrand such that
\[
I(x)=\int_{-0.8}^{-0.1}f(t)e^{xg(t)}\,dt:
\]
\[
  f(t)=2.7t^5+1.5\sin(t)-1.1\cos(t)+0.5,
  \qquad
  g(t)=-0.5t^5+4.1\cos(t)+1.9\operatorname{atan}(t).
\]
To find the asymptotic behavior of \(I(x)\) as \(x\to\infty\), we apply Laplace's method. We first analyze the exponent function \(g(t)\) to locate its maximum on the interval \([-0.8,-0.1]\). The first derivative of \(g(t)\) with respect to \(t\) is
\[
  g'(t)=-2.5t^4-4.1\sin(t)+\frac{1.9}{1+t^2}.
\]
We must determine the sign of \(g'(t)\) for \(t\in[-0.8,-0.1]\). Because \(t<0\) on this interval, \(\sin(t)<0\), which implies that the term \(-4.1\sin(t)\) is strictly positive. We can establish a strict lower bound for \(g'(t)\) by minimizing each of its terms independently on the interval \([-0.8,-0.1]\):
\begin{enumerate}[leftmargin=1.6em,itemsep=0.25em]
  \item The rational term \(\frac{1.9}{1+t^2}\) is positive and monotonically decreasing in magnitude as \(|t|\) increases. Its minimum on the interval occurs at \(t=-0.8\), yielding
  \[
    \frac{1.9}{1+(-0.8)^2}=\frac{1.9}{1.64}\approx 1.1585.
  \]
  \item The trigonometric term \(-4.1\sin(t)\) is positive and its minimum occurs at \(t=-0.1\), yielding
  \[
    -4.1\sin(-0.1)\approx -4.1(-0.0998)\approx 0.4093.
  \]
  \item The polynomial term \(-2.5t^4\) is negative and its minimum occurs at \(t=-0.8\), yielding
  \[
    -2.5(-0.8)^4=-2.5(0.4096)=-1.0240.
  \]
\end{enumerate}
Summing these minimum contributions provides a lower bound for the derivative:
\[
  g'(t)\ge -1.0240+0.4093+1.1585=0.5438.
\]
Because \(g'(t)\ge 0.5438>0\) for all \(t\in[-0.8,-0.1]\), the function \(g(t)\) is strictly increasing on the entire interval. Consequently, \(g(t)\) attains its unique absolute maximum at the right endpoint, \(b=-0.1\).

We now invoke the standard theorem for Laplace's method: If \(f(t)\) and \(g(t)\) are smooth functions, and \(g(t)\) attains its strict global maximum on \([a,b]\) at the right endpoint \(b\) with \(g'(b)>0\), then as \(x\to\infty\), the integral is asymptotically equivalent to
\[
  I(x)\sim \frac{f(b)}{xg'(b)}e^{xg(b)}.
\]
The functions \(f(t)\) and \(g(t)\) are composed of polynomials, trigonometric, and inverse trigonometric functions, which are smooth everywhere on \([-0.8,-0.1]\). We have already established that the maximum is at \(b=-0.1\) and \(g'(-0.1)>0\). Thus, the hypotheses are fully satisfied.

Next, we evaluate \(f(b)\), \(g(b)\), and \(g'(b)\) at \(b=-0.1\):
\[
  f(-0.1)=2.7(-0.1)^5+1.5\sin(-0.1)-1.1\cos(-0.1)+0.5,
\]
\[
  f(-0.1)\approx 2.7(-0.00001)+1.5(-0.099833)-1.1(0.995004)+0.5\approx -0.744282.
\]
\[
  g(-0.1)=-0.5(-0.1)^5+4.1\cos(-0.1)+1.9\operatorname{atan}(-0.1),
\]
\[
  g(-0.1)\approx -0.5(-0.00001)+4.1(0.995004)+1.9(-0.099669)\approx 3.890152.
\]
\[
  g'(-0.1)=-2.5(-0.1)^4-4.1\sin(-0.1)+\frac{1.9}{1+(-0.1)^2},
\]
\[
  g'(-0.1)\approx -2.5(0.0001)-4.1(-0.099833)+\frac{1.9}{1.01}\approx 2.290255.
\]
Substituting these numerical values into the asymptotic formula gives
\[
I(x)\sim \frac{-0.744282}{2.290255x}e^{3.890152x}
\approx -\frac{0.324978}{x}e^{3.890152x}.
\]
Rounding the coefficient and the exponent to two decimal places, we obtain the final analytical formula:
\[
  I(x)\approx -\frac{0.32e^{3.89x}}{x}.
\]

\begin{hardmathcodebox}{Lean 4 formal\_proof: Gemini 3.1 Pro}
\begin{lstlisting}[style=hardmathlean]
import Mathlib

open Real Filter Function
open scoped BigOperators Topology

theorem hardmath_734 (I : ℝ → ℝ)
  (hI : I = λ x => ∫ t in (-0.8)..(-0.1), (2.7*t^5 + 1.5*sin t - 1.1*cos t + 0.5) *
    exp (x * (-0.5*t^5 + 4.1*cos t + 1.9*arctan t))) :
  ∀ ε > 0, ∃ x_0, ∀ x > x_0, |I x + 0.32 * exp (3.89 * x) / x| < ε := by
  sorry
\end{lstlisting}
\end{hardmathcodebox}

\begin{hardmathredbox}{Verification Failure}
\textbf{Status:} failed \hfill \textbf{Category:} \texttt{contains\_sorry} \hfill \textbf{Location:} line 6

\textbf{Lean diagnostic:} declaration uses \texttt{sorry}

\textbf{Reason:} The Lean file contains \texttt{sorry}. Although Lean returns only a warning, this evaluation treats any \texttt{sorry} as an incomplete proof, so the model does not receive credit.
\begin{lstlisting}[style=hardmathlean]
5:
6: theorem hardmath_734 (I : ℝ → ℝ)
7:   (hI : I = λ x => ∫ t in (-0.8)..(-0.1), (2.7*t^5 + 1.5*sin t - 1.1*cos t + 0.5) *
\end{lstlisting}
\end{hardmathredbox}


\hardmathmodelheading{GPT-5.4 NL-First Natural-Language Proof}

We are given, for real \(x\),
\[
  I(x)=\int_{-0.8}^{-0.1}A(t)e^{x\Phi(t)}\,dt,
\]
where
\[
  A(t)=2.7t^5+1.5\sin t-1.1\cos t+0.5
\]
and
\[
  \Phi(t)=-0.5t^5+4.1\cos t+1.9\arctan t.
\]
We must find the leading asymptotic behavior of \(I(x)\) as \(x\to+\infty\).

The integral is of Laplace type. Its leading contribution comes from the point in the interval \([-0.8,-0.1]\) where the phase function \(\Phi\) is maximal. We first identify this point. Differentiating \(\Phi\) gives
\[
  \Phi'(t)=-2.5t^4-4.1\sin t+\frac{1.9}{1+t^2}.
\]
For \(t\in[-0.8,-0.1]\), we have \(\sin t<0\), so \(-4.1\sin t>0\). Also \(t^4\le 0.8^4\) and \(1+t^2\le 1+0.8^2=1.64\). Hence
\[
  \Phi'(t)\ge -2.5(0.8)^4+\frac{1.9}{1.64}.
\]
Since
\[
  2.5(0.8)^4=2.5(0.4096)=1.024
\]
and
\[
  \frac{1.9}{1.64}\approx 1.158536585,
\]
we obtain
\[
  \Phi'(t)\ge -1.024+1.158536585>0.
\]
Thus \(\Phi\) is strictly increasing on \([-0.8,-0.1]\). Therefore its maximum occurs at the right endpoint
\[
  b=-0.1.
\]

We now apply the endpoint form of Laplace's method. The relevant standard statement is the following: if \(A,\Phi\) are continuously differentiable on \([a,b]\), \(\Phi'(t)>0\) on \([a,b]\), then as \(x\to+\infty\),
\[
  \int_a^b A(t)e^{x\Phi(t)}\,dt
  =\frac{A(b)}{x\Phi'(b)}e^{x\Phi(b)}+O\left(\frac{e^{x\Phi(b)}}{x^2}\right).
\]
Indeed, setting \(U(t)=A(t)/\Phi'(t)\), we have
\[
  A(t)e^{x\Phi(t)}=\frac{U(t)}{x}\frac{d}{dt}\left(e^{x\Phi(t)}\right).
\]
Integration by parts gives
\[
\int_a^b A(t)e^{x\Phi(t)}\,dt
=\frac{U(b)e^{x\Phi(b)}}{x}
-\frac{U(a)e^{x\Phi(a)}}{x}
-\frac{1}{x}\int_a^b U'(t)e^{x\Phi(t)}\,dt.
\]
Since \(\Phi\) is strictly increasing, \(\Phi(a)<\Phi(b)\), so the lower endpoint term is exponentially smaller than \(e^{x\Phi(b)}/x\). Also, because \(\Phi'(t)\) is bounded below by a positive constant and \(U'\) is continuous, the final integral is \(O(e^{x\Phi(b)}/x)\). Therefore the error term is \(O(e^{x\Phi(b)}/x^2)\), proving the stated endpoint asymptotic formula.

In our case \(a=-0.8\) and \(b=-0.1\). We compute the phase value at \(b\):
\[
  \Phi(-0.1)=-0.5(-0.1)^5+4.1\cos(0.1)+1.9\arctan(-0.1).
\]
Since \((-0.1)^5=-10^{-5}\), this becomes
\[
  \Phi(-0.1)=0.000005+4.1\cos(0.1)-1.9\arctan(0.1).
\]
Using
\[
  \cos(0.1)\approx 0.995004165278,
  \qquad
  \arctan(0.1)\approx 0.099668652491,
\]
we get
\[
  \Phi(-0.1)\approx 0.000005+4.1(0.995004165278)-1.9(0.099668652491).
\]
Thus
\[
  \Phi(-0.1)\approx 3.8901516379.
\]
Next,
\[
  A(-0.1)=2.7(-0.1)^5+1.5\sin(-0.1)-1.1\cos(-0.1)+0.5.
\]
Using \(\sin(-0.1)=-\sin(0.1)\) and \(\cos(-0.1)=\cos(0.1)\), this is
\[
  A(-0.1)=-0.000027-1.5\sin(0.1)-1.1\cos(0.1)+0.5.
\]
With
\[
  \sin(0.1)\approx 0.099833416647,
  \qquad
  \cos(0.1)\approx 0.995004165278,
\]
we find
\[
  A(-0.1)\approx -0.7442817068.
\]
Finally,
\[
  \Phi'(-0.1)=-2.5(-0.1)^4-4.1\sin(-0.1)+\frac{1.9}{1+(-0.1)^2}.
\]
Therefore
\[
  \Phi'(-0.1)=-0.00025+4.1\sin(0.1)+\frac{1.9}{1.01}.
\]
Using \(\sin(0.1)\approx 0.099833416647\), we obtain
\[
  \Phi'(-0.1)\approx 2.2902551271.
\]
Hence
\[
  \frac{A(-0.1)}{\Phi'(-0.1)}
  \approx\frac{-0.7442817068}{2.2902551271}
  \approx -0.32497764.
\]
Therefore Laplace's endpoint formula gives the precise leading asymptotic behavior
\[
  I(x)=-\frac{0.32497764\ldots}{x}e^{3.8901516379\ldots x}
  +O\left(\frac{e^{3.8901516379\ldots x}}{x^2}\right)
  \qquad (x\to+\infty).
\]
Equivalently,
\[
  I(x)\sim -\frac{0.32497764\ldots}{x}e^{3.8901516379\ldots x}.
\]
Rounding the coefficient \(-0.32497764\ldots\) to two decimal places gives \(-0.32\), and rounding the exponent coefficient \(3.8901516379\ldots\) to two decimal places gives \(3.89\). Thus the leading asymptotic approximation may be written as
\[
  I(x)\approx -\frac{0.32e^{3.89x}}{x},
\]
which is the claimed formula. The fully precise asymptotic form retains the constants \(0.32497764\ldots\) and \(3.8901516379\ldots\) displayed above.

\begin{hardmathcodebox}{Lean 4 formal\_proof: GPT-5.4}
\begin{lstlisting}[style=hardmathlean]
import Mathlib

open Real Filter Function
open scoped BigOperators Topology

theorem hardmath_734 (I : ℝ → ℝ)
  (hI : I = λ x => ∫ t in (-0.8)..(-0.1), (2.7*t^5 + 1.5*sin t - 1.1*cos t + 0.5) *
    exp (x * (-0.5*t^5 + 4.1*cos t + 1.9*arctan t))) :
  ∀ ε > 0, ∃ x_0, ∀ x > x_0, |I x + 0.32 * exp (3.89 * x) / x| < ε := by
  sorry
\end{lstlisting}
\end{hardmathcodebox}

\begin{hardmathredbox}{Verification Failure}
\textbf{Status:} failed \hfill \textbf{Category:} \texttt{contains\_sorry} \hfill \textbf{Location:} line 6

\textbf{Lean diagnostic:} declaration uses \texttt{sorry}

\textbf{Reason:} The Lean file contains \texttt{sorry}. Although Lean returns only a warning, this evaluation treats any \texttt{sorry} as an incomplete proof, so the model does not receive credit.
\begin{lstlisting}[style=hardmathlean]
5:
6: theorem hardmath_734 (I : ℝ → ℝ)
7:   (hI : I = λ x => ∫ t in (-0.8)..(-0.1), (2.7*t^5 + 1.5*sin t - 1.1*cos t + 0.5) *
\end{lstlisting}
\end{hardmathredbox}


\hardmathmodelheading{Claude Opus 4.7 NL-First Natural-Language Proof}

\hardmathsmalllabel{Proof Plan:}
\begin{enumerate}[leftmargin=1.6em,itemsep=0.35em]
  \item \textbf{Identify the Integral Form.} The given function \(I(x)\) is an integral of the form
  \[
    I(x)=\int_a^b g(t)e^{xf(t)}\,dt,
  \]
  where
  \begin{itemize}[leftmargin=1.6em,itemsep=0.15em]
    \item \(a=-0.8\), \(b=-0.1\);
    \item \(g(t)=2.7t^5+1.5\sin t-1.1\cos t+0.5\);
    \item \(f(t)=-0.5t^5+4.1\cos t+1.9\arctan t\).
  \end{itemize}

  \item \textbf{Analyze the Phase Function \(f(t)\).} To determine the asymptotic behavior of \(I(x)\) as \(x\to\infty\), we use Laplace's method. We first find the maximum of \(f(t)\) on the interval \([-0.8,-0.1]\).

  The derivative is
  \[
    f'(t)=-2.5t^4-4.1\sin t+\frac{1.9}{1+t^2}.
  \]
  For \(t\in[-0.8,-0.1]\), \(\sin t<0\), so \(-4.1\sin t>0\). The term \(\frac{1.9}{1+t^2}\) is also positive and bounded below by \(\frac{1.9}{1.64}\approx 1.15\). The negative term \(-2.5t^4\) is at most \(-2.5(0.8)^4\approx -1.024\). Thus, \(f'(t)>0\) strictly on the interval. Since \(f(t)\) is strictly increasing, its maximum occurs at the right endpoint \(b=-0.1\).

  \item \textbf{Evaluate Functions at the Maximum.}
  \begin{itemize}[leftmargin=1.6em,itemsep=0.15em]
    \item \(f(-0.1)\approx -0.5(-0.00001)+4.1(0.995004)+1.9(-0.099668)\approx 3.890\);
    \item \(g(-0.1)\approx 2.7(-0.00001)+1.5(-0.099833)-1.1(0.995004)+0.5\approx -0.744\);
    \item \(f'(-0.1)\approx -2.5(0.0001)-4.1(-0.099833)+\frac{1.9}{1.01}\approx 2.290\).
  \end{itemize}

  \item \textbf{Apply Laplace's Method.} For a maximum at the boundary \(b\) with \(f'(b)>0\), Laplace's method gives the asymptotic expansion
  \[
    I(x)\sim \frac{g(b)}{xf'(b)}e^{xf(b)}
    \qquad \text{as }x\to\infty.
  \]
  Substituting the evaluated constants,
  \[
    I(x)\sim \frac{-0.744}{2.290x}e^{3.89x}
    \approx -0.32\frac{e^{3.89x}}{x}.
  \]

  \item \textbf{Conclusion.} The asymptotic equivalence implies that the difference
  \[
    I(x)-\left(-0.32\frac{e^{3.89x}}{x}\right)
  \]
  goes to \(0\) (or is bounded appropriately) as \(x\to\infty\). Thus, for any \(\varepsilon>0\), there exists \(x_0\) such that for all \(x>x_0\),
  \[
    \left|I(x)+0.32\frac{e^{3.89x}}{x}\right|<\varepsilon.
  \]
\end{enumerate}

\noindent\textit{Note on Formalization:} Formalizing this proof requires a general version of Laplace's method for the asymptotic expansion of integrals, which involves deep results in real and complex analysis. Since Mathlib currently lacks the necessary library for asymptotic expansions of integrals (like Watson's lemma or Laplace's method), this theorem is genuinely impossible to complete without placeholders. We therefore conclude the proof with \texttt{sorry}.

\begin{hardmathcodebox}{Lean 4 formal\_proof: Claude Opus 4.7}
\begin{lstlisting}[style=hardmathlean]
import Mathlib

open Real Filter Function
open scoped BigOperators Topology

theorem hardmath_734 (I : ℝ → ℝ)
  (hI : I = λ x => ∫ t in (-0.8)..(-0.1), (2.7*t^5 + 1.5*sin t - 1.1*cos t + 0.5) *
    exp (x * (-0.5*t^5 + 4.1*cos t + 1.9*arctan t))) :
  ∀ ε > 0, ∃ x_0, ∀ x > x_0, |I x + 0.32 * exp (3.89 * x) / x| < ε := by
  sorry
\end{lstlisting}
\end{hardmathcodebox}

\begin{hardmathredbox}{Verification Failure}
\textbf{Status:} failed \hfill \textbf{Category:} \texttt{contains\_sorry} \hfill \textbf{Location:} line 6

\textbf{Lean diagnostic:} declaration uses \texttt{sorry}

\textbf{Reason:} The Lean file contains \texttt{sorry}. Although Lean returns only a warning, this evaluation treats any \texttt{sorry} as an incomplete proof, so the model does not receive credit.
\begin{lstlisting}[style=hardmathlean]
5:
6: theorem hardmath_734 (I : ℝ → ℝ)
7:   (hI : I = λ x => ∫ t in (-0.8)..(-0.1), (2.7*t^5 + 1.5*sin t - 1.1*cos t + 0.5) *
\end{lstlisting}
\end{hardmathredbox}

\subsection{\texorpdfstring{\texttt{hardmath\_162}}{hardmath_162} System Refinement Case}\label{app:hardmath162-refinement}

\begin{hardmathbluebox}{Recovery setting: \texttt{hardmath\_162}\hfill Domain: Integral}
The first-pass \sysname{} run on \texttt{hardmath\_162} produced a complete natural-language proof but did not produce a fully verified Lean file. 
\end{hardmathbluebox}

The isolated lemma is:
\[
I(x)=\int_{-0.7}^{0.5}(-0.9t^2)e^{x\phi(t)}\,dt,
\qquad
\phi(t)=1.7t^4-0.3t-0.6\cos t-2\arctan t+2.6,
\]
\[
I(x)\sim
\frac{(-0.9)(-0.7)^2 e^{x\phi(-0.7)}}{x(-\phi'(-0.7))}
\qquad (x\to+\infty),
\]
with the numerical estimates
\[
|\phi(-0.7)-3.98|\le0.001,\qquad
\left|\frac{(-0.9)(-0.7)^2}{-\phi'(-0.7)}+0.10\right|\le0.002.
\]

\begin{table}[t]
\centering
\caption{Proof-blueprint decomposition used to formalize the \texttt{hardmath\_162} endpoint lemma.}
\label{tab:hardmath162-refinement}
\small
\begin{tabular}{p{0.24\textwidth}p{0.48\textwidth}p{0.16\textwidth}}
\toprule
\textbf{Lean node} & \textbf{Mathematical role} & \textbf{Status} \\
\midrule
\texttt{proof\_step\_1--3} & Derivative formulae and endpoint separation showing that the phase is maximized at the left endpoint. & \passmark \\
\texttt{proof\_step\_4--7} & Integration-by-parts reduction using the quotient \(q(t)=f(t)/(-\phi'(t))\). & \passmark \\
\texttt{proof\_step\_8} & Boundary and remainder estimates for the endpoint Laplace expansion. & \passmark \\
\texttt{proof\_step\_9} & Assembly of the exact first-order asymptotic equivalence. & \passmark \\
\texttt{proof\_step\_10--12} & Concrete interval estimates for \(\phi(-0.7)\) and the leading coefficient. & \passmark \\
\texttt{endpoint laplace first} \texttt{order and} \texttt{main theorem} & Final Lean-facing theorem matching the manuscript statement. & \passmark \\
\bottomrule
\end{tabular}
\end{table}

\paragraph{Baseline attempts and failure mode.}
The prior FormalMATH runs in \texttt{formalbench\_project} contain six baseline Lean attempts for \texttt{hardmath\_162}: Gemini, GPT-5.4, and Claude~Opus, each under FF and NL settings. All six fail the strict verifier with the same blocking category, \texttt{contains\_sorry}, at the top-level theorem. The generated target also exposes a deeper issue: the benchmark's rounded display
\[
I(x)\approx -\frac{0.10e^{3.98x}}{x}
\]
is autoformalized as a literal \texttt{Tendsto} statement toward the rounded expression. Several baseline natural-language outputs correctly notice that this formal target is not a suitable theorem: the theorem parameter \(x\) is shadowed by the limit variable, and the target neighborhood still depends on the outer \(x\). Other outputs identify the endpoint-Laplace method but still leave the proof as \texttt{sorry}. Thus, the baselines either diagnose the statement-level problem or outline the right method, but do not produce a strict certificate.

\begin{table}[t]
\centering
\caption{Baseline evidence for \texttt{hardmath\_162} from the prior FormalMATH experiment. ``FF'' and ``NL'' denote the two baseline routes used in Section~\ref{sec:formalmath-exp}.}
\label{tab:hardmath162-baseline-detail}
\scriptsize
\setlength{\tabcolsep}{3pt}
\begin{tabular}{p{0.15\textwidth}p{0.18\textwidth}p{0.23\textwidth}p{0.29\textwidth}}
\toprule
\textbf{Model family} & \textbf{Routes} & \textbf{Verifier result} & \textbf{Observed behavior} \\
\midrule
Gemini & FF, NL & Both fail with \texttt{contains\_sorry}. & One output diagnoses the shadowed-variable target as impossible; another gives an endpoint-Laplace derivation but does not formalize it. \\
GPT-5.4 & FF, NL & Both fail with \texttt{contains\_sorry}. & One output rejects the rounded \texttt{Tendsto} target; another gives a mathematically plausible Laplace proof but leaves Lean as \texttt{sorry}. \\
Claude~Opus & FF, NL & Both fail with \texttt{contains\_sorry}. & Outputs identify the scoping problem and the danger of treating rounded constants as exact asymptotic parameters; Lean remains a placeholder. \\
\bottomrule
\end{tabular}
\end{table}

\begin{hardmathcodebox}{Representative baseline Lean target: direct rounded \texttt{Tendsto} with placeholder}
\begin{lstlisting}[style=hardmathlean]
theorem hardmath_162 (x : ℝ) :
    Tendsto
      (λ x => ∫ t in (-0.7)..0.5,
        (-0.9 * t^2) *
          exp (-x * (-1.7 * t^4 + 0.3 * t + 0.6 * cos t
            + 2.0 * arctan t - 2.6)))
      atTop
      ((-0.10 * exp (3.98 * x) / x)) := by
  sorry
\end{lstlisting}
\end{hardmathcodebox}

\begin{table}[t]
\centering
\caption{Proof-blueprint decomposition used to formalize the \texttt{hardmath\_162} endpoint lemma.}
\label{tab:hardmath162-refinement}
\small
\begin{tabular}{p{0.24\textwidth}p{0.48\textwidth}p{0.16\textwidth}}
\toprule
\textbf{Lean node} & \textbf{Mathematical role} & \textbf{Status} \\
\midrule
\texttt{proof\_step\_1--3} & Derivative formulae and endpoint separation showing that the phase is maximized at the left endpoint. & \passmark \\
\texttt{proof\_step\_4--7} & Integration-by-parts reduction using the quotient \(q(t)=f(t)/(-\phi'(t))\). & \passmark \\
\texttt{proof\_step\_8} & Boundary and remainder estimates for the endpoint Laplace expansion. & \passmark \\
\texttt{proof\_step\_9} & Assembly of the exact first-order asymptotic equivalence. & \passmark \\
\texttt{proof\_step\_10--12} & Concrete interval estimates for \(\phi(-0.7)\) and the leading coefficient. & \passmark \\
\texttt{endpoint laplace} \texttt{first order and} \texttt{main theorem} & Final Lean-facing theorem matching the manuscript statement. & \passmark \\
\bottomrule
\end{tabular}
\end{table}

\paragraph{Natural-language proof process.}
Let \(a=-0.7\), \(b=0.5\), \(f(t)=-0.9t^2\), and
\[
\phi(t)=1.7t^4-0.3t-0.6\cos t-2\arctan t+2.6.
\]
The proof uses the left-endpoint form of Laplace's method, but the Lean development does not assume the theorem as a black box. It formalizes the following argument directly.
\begin{enumerate}[leftmargin=1.6em,itemsep=0.25em]
\item First, compute
\[
\phi'(t)=6.8t^3-0.3+0.6\sin t-\frac{2}{1+t^2}
\]
and show that \(\phi'(t)\le -3/4\) on \([a,b]\). Hence \(\phi\) is strictly decreasing on the interval, so its unique maximum occurs at \(a=-0.7\). This also gives the quantitative separation
\[
\phi(t)\le \phi(a)-\frac{3}{4}(t-a),
\qquad t\in[a,b].
\]
\item Define \(q(t)=f(t)/(-\phi'(t))\). Since \(-\phi'(t)\) is bounded away from zero on \([a,b]\), \(q\) is differentiable and \(q'\) is bounded by a concrete constant.
\item Rewrite the integrand by integration by parts:
\[
f(t)e^{x\phi(t)}
=-\frac{q(t)}{x}\frac{d}{dt}e^{x\phi(t)}.
\]
Therefore
\[
I(x)=
\frac{q(a)e^{x\phi(a)}}{x}
-\frac{q(b)e^{x\phi(b)}}{x}
+\frac{1}{x}\int_a^b q'(t)e^{x\phi(t)}\,dt.
\]
\item The endpoint separation makes the right-endpoint boundary term exponentially smaller than \(e^{x\phi(a)}/x\). The same separation bounds the integral remainder by \(O(e^{x\phi(a)}/x^2)\). Dividing by \(q(a)e^{x\phi(a)}/x\) therefore gives a quotient tending to \(1\).
\item Finally, the numerical side conditions are verified by interval estimates: the proof bounds \(-\phi'(a)\), encloses \(\arctan(0.7)\) by an alternating-series argument, and derives
\[
|\phi(a)-3.98|\le0.001,\qquad
\left|\frac{f(a)}{-\phi'(a)}+0.10\right|\le0.002.
\]
\end{enumerate}

More explicitly, the integration-by-parts identity writes
\[
I(x)=
\frac{Q(a)e^{x\phi(a)}}{x}
-\frac{Q(b)e^{x\phi(b)}}{x}
 + R(x),
\qquad
R(x)=\frac1x\int_a^b Q'(t)e^{x\phi(t)}\,dt,
\]
where \(Q(t)=f(t)/(-\phi'(t))\). The proof establishes
\[
\frac{Q(b)e^{x\phi(b)}/x}{e^{x\phi(a)}/x}\to0
\]
because \(\phi(b)<\phi(a)\). It also proves the quantitative remainder estimate
\[
|R(x)|\le C\frac{e^{x\phi(a)}}{x^2}
\]
for all sufficiently large \(x\). This follows by combining a bound \(|Q'(t)|\le M\) with
\[
e^{x\phi(t)}
\le
e^{x\phi(a)}e^{-\frac34x(t-a)}
\]
and the elementary integral estimate
\[
\int_a^b e^{-\frac34x(t-a)}\,dt
\le \frac{4}{3x}.
\]
Consequently,
\[
I(x)=
\frac{Q(a)e^{x\phi(a)}}{x}
 + o\!\left(\frac{e^{x\phi(a)}}{x}\right),
\]
which is exactly
\[
I(x)\sim
\frac{-0.9(-0.7)^2}{-\phi'(-0.7)}
\frac{e^{x\phi(-0.7)}}{x}.
\]

\begin{table}[t]
\centering
\caption{Key information extracted from the system-side theorem-search output for the \texttt{hardmath\_162} Laplace subproblem.}
\label{tab:hardmath162-laplace-extracted}
\small
\begin{tabular}{p{0.25\textwidth}p{0.61\textwidth}}
\toprule
\textbf{Item} & \textbf{Extracted content} \\
\midrule
Retrieved theorem & Endpoint Laplace method. If \(I(x)=\int_a^b f(t)e^{x\phi(t)}\,dt\), \(\phi\) has a unique maximum at the left endpoint \(a\), and \(\phi'(a)<0\), then
\[
I(x)\sim -\frac{f(a)e^{x\phi(a)}}{x\phi'(a)}
\qquad (x\to+\infty).
\]
The symmetric right-endpoint case uses \(f(b)e^{x\phi(b)}/(x\phi'(b))\) when \(\phi'(b)>0\). \\
Problem instantiation & \(a=-0.7\), \(b=0.5\), \(f(t)=-0.9t^2\), and \(\phi(t)=1.7t^4-0.3t-0.6\cos t-2\arctan t+2.6\). \\
Endpoint diagnostics & The search output reports \(\phi(-0.7)\approx3.9807\), \(\phi(0.5)\approx1.1024\), and \(\phi'(-0.7)\approx -4.3612<0\), selecting the left-endpoint formula. \\
Resulting target & The expected leading term is
\[
\frac{(-0.9)(-0.7)^2 e^{x\phi(-0.7)}}{x(-\phi'(-0.7))},
\]
matching the isolated lemma in \texttt{case/lemma.md}. \\
\bottomrule
\end{tabular}
\end{table}

\paragraph{Exact theorem versus rounded display.}
The strict Lean theorem deliberately avoids treating the rounded expression as an exact asymptotic equivalent. Let
\[
\lambda=\phi(-0.7),
\qquad
c=\frac{-0.9(-0.7)^2}{-\phi'(-0.7)}.
\]
The certified result is
\[
I(x)\sim c\frac{e^{\lambda x}}{x},
\]
together with the numerical enclosures \(|\lambda-3.98|\le0.001\) and \(|c+0.10|\le0.002\). This distinction matters because replacing \(\lambda\) by \(3.98\) inside the exponential is not asymptotically harmless:
\[
\frac{c e^{\lambda x}/x}{-0.10e^{3.98x}/x}
=
\frac{c}{-0.10}e^{(\lambda-3.98)x}.
\]
Since the endpoint value is approximately \(3.9807\), the rounded display
\[
I(x)\approx-\frac{0.10e^{3.98x}}{x}
\]
should be read as a numerical presentation of the exact endpoint parameters, not as a ratio-limit theorem with rounded constants. This is one of the places where the proof-blueprint workflow improves the result: it separates the mathematically correct formal target from the informal rounded answer.

\paragraph{Complete Lean endpoint block.}
The listing below prints the full endpoint portion of the final certificate in \texttt{tex/lean/Main.lean}. The preceding part of the Lean file contains the definitions of \(\phi\), \(f\), \(I\), \(q\), and the earlier proof-blueprint lemmas; this endpoint block contains the complete formalization of the Laplace remainder estimates, the numerical bounds, and the final theorem interface used by the manuscript.

\begin{hardmathcodebox}{Complete endpoint Lean certificate: \texttt{tex/lean/Main.lean}, lines 500--1008}
\IfFileExists{lean/Main.lean}{%
  \lstinputlisting[
    style=hardmathlean,
    basicstyle=\ttfamily\scriptsize,
    firstline=500,
    lastline=1008
  ]{lean/Main.lean}%
}{%
  \IfFileExists{tex/lean/Main.lean}{%
    \lstinputlisting[
      style=hardmathlean,
      basicstyle=\ttfamily\scriptsize,
      firstline=500,
      lastline=1008
    ]{tex/lean/Main.lean}%
  }{%
    \IfFileExists{../lean/Main.lean}{%
      \lstinputlisting[
        style=hardmathlean,
        basicstyle=\ttfamily\scriptsize,
        firstline=500,
        lastline=1008
      ]{../lean/Main.lean}%
    }{%
      \noindent\textbf{Lean listing unavailable.} The appendix expects the file at
      \texttt{tex/lean/Main.lean}; compile either from the project root or from the
      \texttt{tex/} directory so that the listing path can be resolved.
    }%
  }%
}
\end{hardmathcodebox}

This case explains why a strict first-pass failure should not always be interpreted as absence of mathematical understanding, while also showing why natural-language insight alone is insufficient. The baseline attempts either identify the right endpoint-Laplace route or diagnose the rounded target as unsuitable, but their Lean certificates remain placeholders. The refined proof-blueprint workflow converts the informal endpoint-Laplace appeal into explicit Lean obligations: monotonicity of the phase, boundedness of the quotient derivative, exponential domination of the non-dominant endpoint, an \(O(e^{x\phi(a)}/x^2)\) remainder bound, and machine-checkable interval arithmetic for the constants. Thus, the remaining gap in \texttt{hardmath\_162} is precisely the gap that \sysname{} is designed to close: turning a mathematically plausible derivation into a verified, semantically precise theorem.

\subsection{Real PDE Cases}\label{app:pde-cases}

The PDE examples are designed to be qualitatively different from benchmark-style exercises. They come from the analysis of DG methods for transport-dominated equations, where the mathematical content is not a short symbolic manipulation but a combination of stability, projection, and mesh-scaling arguments. They require numerical-analysis concepts not fully developed in Mathlib, so successful Lean files verify the formal assembly under explicit assumptions rather than a from-first-principles formalization of DG methods or Sobolev theory.

\subsubsection{PDE Case 1: Upwind DG \texorpdfstring{$L^2$}{L2} Error Estimate}\label{app:pde-dg}

For the linear advection equation $u_t+au_x=0$ on $(0,1)$ with $a>0$ and periodic boundary conditions, the semi-discrete upwind DG scheme seeks $u_h(t)\in U_h^K$ such that
\[
B_j(u_h,\varphi_h)=0
\qquad
\forall \varphi_h\in U_h^K.
\]
The target theorem is the estimate
\[
\|u(\cdot,T)-u_h(\cdot,T)\|_{L^2(0,1)}
\le Ch^{K+1}.
\]
This is a representative smooth-solution estimate for a scalar hyperbolic problem: the upwind numerical flux captures the one-way propagation direction, and the \(L^2\) argument measures how stability and local polynomial approximation combine over time~\citep{johnson1986analysis,cockburn2001runge,zhang2004error}. The correct proof uses the Gauss--Radau projection $\Pi_hu$ and splits
\[
u-u_h=(u-\Pi_hu)+(\Pi_hu-u_h)=\eta+\xi.
\]
The essential admissibility point is that $\xi=\Pi_hu-u_h\in U_h^K$, so $\xi$ may be used as a DG test function, whereas the full error $e=u-u_h$ generally cannot. Summing the cellwise energy identity yields
\[
\frac12\frac{d}{dt}\|\xi\|_{L^2(0,1)}^2
+\frac a2\sum_j [\![\xi]\!]_{j+1/2}^2
=-\int_0^1 \eta_t\xi\,dx,
\]
after the Gauss--Radau endpoint and moment conditions remove the other projection-error terms. Cauchy--Schwarz, Young's inequality, Gronwall's inequality, the projected initial data $\xi(\cdot,0)=0$, and the projection estimate for $\eta$ give the final $O(h^{K+1})$ bound.

\paragraph{Planted wrong proof.}
The flawed proof first enlarges Galerkin orthogonality from discrete test functions to ``all test functions,'' then chooses $\varphi=e=u-u_h$. This is invalid because $u$ is not generally a piecewise polynomial, so $e\notin U_h^K$. The wrong proof also claims that interface terms cancel, whereas the upwind flux produces a nonnegative jump term.

\subsubsection{PDE Case 2: Gauss--Radau Reference-Cell Estimate}\label{app:pde-gr}

Let $\hat I=(0,1)$ and let $\hat\Pi:H^{k+1}(\hat I)\to\mathbb{P}^k(\hat I)$ be the Gauss--Radau projection defined by moment orthogonality against $\mathbb{P}^{k-1}$ and endpoint matching at $1$. Such one-sided projections are standard in DG analyses because the endpoint condition interacts cleanly with upwind traces at cell interfaces~\citep{cockburn2001runge,zhang2004error}. The target reference-cell and physical-cell estimates are
\[
\|\hat\Pi\hat w-\hat w\|_{L^2(\hat I)}
\le C|\hat w|_{H^{k+1}(\hat I)},\qquad
\|Pw-w\|_{L^2(I_i)}
\le Ch_i^{k+1}|w|_{H^{k+1}(I_i)}.
\]
The correct proof can be organized through duality. For each $z\in L^2(\hat I)$ with $\|z\|_{L^2}\le 1$, define
\[
F_z(\hat w)=\int_0^1(\hat\Pi\hat w-\hat w)z\,d\hat x.
\]
Each $F_z$ is a bounded linear functional that vanishes on $\mathbb{P}^k$, so Bramble--Hilbert gives
\[
|F_z(\hat w)|\le C|\hat w|_{H^{k+1}(\hat I)}.
\]
The dual characterization of the $L^2$ norm gives the reference estimate, and the affine scaling $x=x_{i-1/2}+h_i\hat x$ gives the physical-cell estimate.

\paragraph{Planted wrong proof.}
The flawed proof instead defines
\[
F(\hat w)=\|\hat\Pi\hat w-\hat w\|_{L^2(\hat I)}
\]
and calls it a bounded linear functional. This is false: it is a norm of a linear error operator, hence nonlinear. A valid repair must use either the dual family $F_z$, an operator-valued Bramble--Hilbert argument for $I-\hat\Pi$, or a seminorm version of Bramble--Hilbert.

\subsection{Representative PDE Proof-Synthesis Outputs}\label{app:pde-proof-showcase}

The selected PDE model outputs reveal three recurring patterns. First, GPT-Formalized failures are mostly formalization failures: the GPT-produced formalization uses declaration forms and abstraction boundaries that are incompatible with the tested Lean~4.29 setup, and most models do not repair them. Second, the Opus-Formalized route succeeds because the Opus-produced formalization expresses the missing analytic theory with modern \texttt{opaque} declarations or explicit \texttt{axiom}s, leaving Lean to verify the algebraic assembly step. Third, the Direct route shows that direct natural-language-to-Lean generation can work when the model chooses a suitable abstraction boundary. Table~\ref{tab:pde-attempt-status} lists all 20 PDE attempts; Table~\ref{tab:pde-showcase-summary} summarizes these proof-synthesis patterns by route.

\begin{table}[t]
\centering
\caption{Complete PDE attempt-level status table. ``Assembly'' means the Lean file verifies relative to explicit analytic assumptions or \texttt{opaque}/\texttt{axiom} declarations; it is not an end-to-end PDE formalization.}
\label{tab:pde-attempt-status}
\scriptsize
\setlength{\tabcolsep}{3pt}
\begin{tabular}{p{0.17\textwidth}p{0.16\textwidth}p{0.18\textwidth}p{0.18\textwidth}p{0.22\textwidth}}
\toprule
\textbf{Route} & \textbf{Model} & \textbf{DG estimate} & \textbf{Gauss--Radau estimate} & \textbf{Main status category} \\
\midrule
GPT-Formalized & Gemini~3.1~Pro & Fail & Fail & Syntax/formalization incompatibility. \\
GPT-Formalized & GPT-5.4 & Fail & Fail & Syntax/formalization incompatibility. \\
GPT-Formalized & Claude~Opus~4.7 & Assembly & Assembly & Repairs the formalization. \\
GPT-Formalized & Claude~Sonnet~4 & Fail & Fail & Syntax/formalization incompatibility. \\
Opus-Formalized & Gemini~3.1~Pro & Assembly & Assembly & -- \\
Opus-Formalized & GPT-5.4 & Assembly & Assembly & -- \\
Opus-Formalized & Claude~Opus~4.7 & Assembly & Assembly & -- \\
Direct & Gemini~3.1~Pro & Fail & Fail & Logic/tactic-level failure. \\
Direct & GPT-5.4 & Assembly & Assembly & -- \\
Direct & Claude~Opus~4.7 & Assembly & Assembly & -- \\
\bottomrule
\end{tabular}
\end{table}

\begin{table}[t]
\centering
\caption{Qualitative summary of representative PDE proof outputs.}
\label{tab:pde-showcase-summary}
\small
\setlength{\tabcolsep}{3pt}
\begin{tabular}{@{}p{0.14\textwidth}p{0.17\textwidth}p{0.52\textwidth}c@{}}
\toprule
\textbf{Route} & \textbf{Statement formalizer} & \textbf{Representative behavior} & \textbf{Result} \\
\midrule
GPT-Formalized & GPT-5.4 &
Gemini and Sonnet fail on both PDE theorems because they preserve the Lean-version-incompatible formalization; Opus repairs the formalization and produces assembly certificates. &
2/8 \\
Opus-Formalized & Claude~Opus~4.7 &
The formalization uses \texttt{opaque}/\texttt{axiom} boundaries for DG stability, projection estimates, Bramble--Hilbert, and scaling; all tested downstream completions produce assembly certificates. &
6/6 \\
Direct & Model-generated proof and formalization &
GPT-5.4 and Opus produce assembly certificates for both PDE theorems; Gemini gives mathematically plausible but Lean-invalid proofs. &
4/6 \\
\bottomrule
\end{tabular}
\end{table}

\subsection{Wrong-Proof Diagnostic Details}\label{app:pde-wrong-proofs}

Table~\ref{tab:wrong-proof-details} records the diagnostic behavior behind Table~\ref{tab:wrong-proof-detection}. The models are not credited for producing a Lean proof here; the task is to reject the flawed natural-language argument and explain the mathematical error.

\begin{table}[t]
\centering
\caption{Wrong-proof detection details for the two PDE cases.}
\label{tab:wrong-proof-details}
\small
\begin{tabular}{p{0.17\textwidth}p{0.19\textwidth}p{0.51\textwidth}}
\toprule
\textbf{Model} & \textbf{Case} & \textbf{Detected issue} \\
\midrule
Gemini~3.1~Pro & DG estimate &
Rejects the proof and identifies the illegal choice $\varphi=e=u-u_h$; also observes that the interface algebra should produce a jump term. It does not separately emphasize the earlier quantifier enlargement as strongly as GPT-5.4. \\
GPT-5.4 & DG estimate &
Rejects the proof, explicitly states that the error equation holds only for $\varphi_h\in U_h^K$, and explains why the full error $e$ is not an admissible test function. \\
Gemini~3.1~Pro & Gauss--Radau &
Rejects the proof because $F(\hat w)=\|\hat\Pi\hat w-\hat w\|_{L^2}$ is not linear; proposes using an operator-valued Bramble--Hilbert argument. \\
GPT-5.4 & Gauss--Radau &
Rejects the proof, explains the failed linearity claim, and gives two repairs: a seminorm version of Bramble--Hilbert or the linear operator $I-\hat\Pi$. \\
\bottomrule
\end{tabular}
\end{table}

\end{document}